%% file: main.tex
\documentclass[twoside,final,1p,times]{elsarticle}
\usepackage{amsmath}
\usepackage{subfigure}
\usepackage{multirow}
\usepackage{url}
\usepackage{fancyhdr}
\usepackage{color}
\journal{Data Intelligence}

\newcommand\shorttitle{\textcolor[RGB]{20,180,190}{Data Intelligence}}
\usepackage{algorithm}
\usepackage{algorithmic}
\usepackage{amssymb}   
\usepackage{booktabs}
\usepackage{multirow}
\usepackage{colortbl}  
\definecolor{tab_others}{RGB}{235, 235, 235}
\definecolor{tab_ours}{RGB}{225, 235, 246}
\usepackage{rotating}
\def \pzo {\phantom{0}} 
\newcommand{\corrmark}{\textsuperscript{\dag}}

\usepackage{soul}
\sethlcolor{yellow!35}
\soulregister\cite7 
\usepackage[colorlinks,linkcolor=blue,citecolor=blue]{hyperref}
\begin{document}
\begin{frontmatter}
\title{Collaborative Reconstruction and Repair for Multi-class Industrial Anomaly Detection}

\author[1,2]{Qishan Wang}
\author[3]{Haofeng Wang\corrmark}
\author[4]{Shuyong Gao}
\author[5]{Jia Guo}
\author[2]{Li Xiong}
\author[2]{Jiaqi Li}
\author[2]{Dengxuan Bai}
\author[1,4]{Wenqiang Zhang\corrmark}

\address[1]{College of Intelligent Robotics and Advanced Manufacturing, Fudan University}
\address[2]{College of Physics and Electromechanical Engineering, Hexi University}
\address[3]{College of Design and Innovation, Tongji University}
\address[4]{Shanghai Key Lab of Intelligent Information Processing, College of Computer Science and Artificial Intelligence, Fudan university}
\address[5]{School of Biomedical Engineering, Tsinghua University}

\cortext[cor1]{Corresponding author: Haofeng Wang (Email: haofen.wang@tongji.edu.cn; ORCID:0000-0003-3018-3824) and Wenqiang Zhang (Email: wqzhang@fudan.edu.cn; ORCID:0000-0002-3339-8751)}

\input{./sec/0_abstract}

\end{frontmatter}

\input{./sec/1_intro}

\input{./sec/2_related_work}

\input{./sec/3_method}

\input{./sec/4_experiments}

\input{./sec/5_conclusion}

\section*{Author Contributions}

\begin{sloppypar}
    Qishan Wang(qswang20@fudan.edu.cn; ORCID: 0000-0003-3463-9040), Wenqiang Zhang (wqzhang@fudan.edu.cn; ORCID: 0000-0002-3339-8751) and Jia Guo(j-g24@mails.tsinghua.edu.cn; ORCID:0000-0002-4449-6867) conceived of the presented idea and designed the framework of CRR. Qishan Wang wrote the manuscript with the help of Haofen Wang(haofen.wang@tongji.edu.cn; ORCID: 0000-0003-3018-3824) and Shuyun Gao(sygao18@fudan.edu.cn; ORCID: 0000-0002-8992-0756). Li Xiong(xl2025@hxu.edu.cn; ORCID: 0000-0003-4615-8367), Jiaqi Li(lijq@hxu.edu.cn; ORCID:  0000-0001-7939-0360) and Dengxuan Bai(baidengxuan@hxu.edu.cn; ORCID: 0000-0002-1359-4819)  provided critical feedback and helped shape the research and analysis.
\end{sloppypar}

\section*{Acknowledgements}
This work was supported in part by the National Natural Science Foundation of China under Grants 62576109, 62072112, and 62461022; in part by the Hexi University President’s Fund for Young Scientists Research Project under Grant QN202204; and in part by the Gansu Provincial Education Scientific and Technological Innovation Project under Grant 2023A-130.

\bibliographystyle{elsarticle-num}
\bibliography{main}

\section*{Author Biography}
\textbf{Qishan Wang} received his Ph.D. degree from Fudan University, Shanghai, China, in 2025. He is currently a faculty member at Hexi University, Zhangye, China. His research interests include industrial anomaly detection, defect detection, and computer vision.

\clearpage
\input{./sec/6_appendix}
\end{document}

%% file: sec/0_abstract.tex
\begin{abstract}
Industrial anomaly detection is a challenging open-set task that aims to identify unknown anomalous patterns deviating from normal data distribution. To avoid the significant memory consumption and limited generalizability brought by building separate models per class, we focus on developing a unified framework for multi-class anomaly detection.
However, under this challenging setting, conventional reconstruction-based networks often suffer from an identity mapping problem, where they directly replicate input features regardless of whether they are normal or anomalous, resulting in detection failures.
To address this issue, this study proposes a novel framework termed Collaborative Reconstruction and Repair (CRR), which transforms the reconstruction to repairation. 
First, we optimize the decoder to reconstruct normal samples while repairing synthesized anomalies.
Consequently, it generates distinct representations for anomalous regions and similar representations for normal areas compared to the encoder's output.
Second, we implement feature-level random masking to ensure that the representations from decoder contain sufficient local information. 
Finally, to minimize detection errors arising from the discrepancies between feature representations from the encoder and decoder, we train a segmentation network supervised by synthetic anomaly masks, thereby enhancing localization performance.
Extensive experiments on industrial datasets that CRR effectively mitigates the identity mapping issue and achieves state-of-the-art performance in multi-class industrial anomaly detection.
\end{abstract}

\begin{keyword}
Computer Vision; Feature Reconstruction; Image Repair; Multi-class Industrial Anomaly Detection; Defect Detection
\end{keyword}


%% file: sec/1_intro.tex
\section{INTRODUCTION}
\label{sec:introduction}
\begin{figure}[!t]
\centering{\includegraphics[width=\columnwidth]{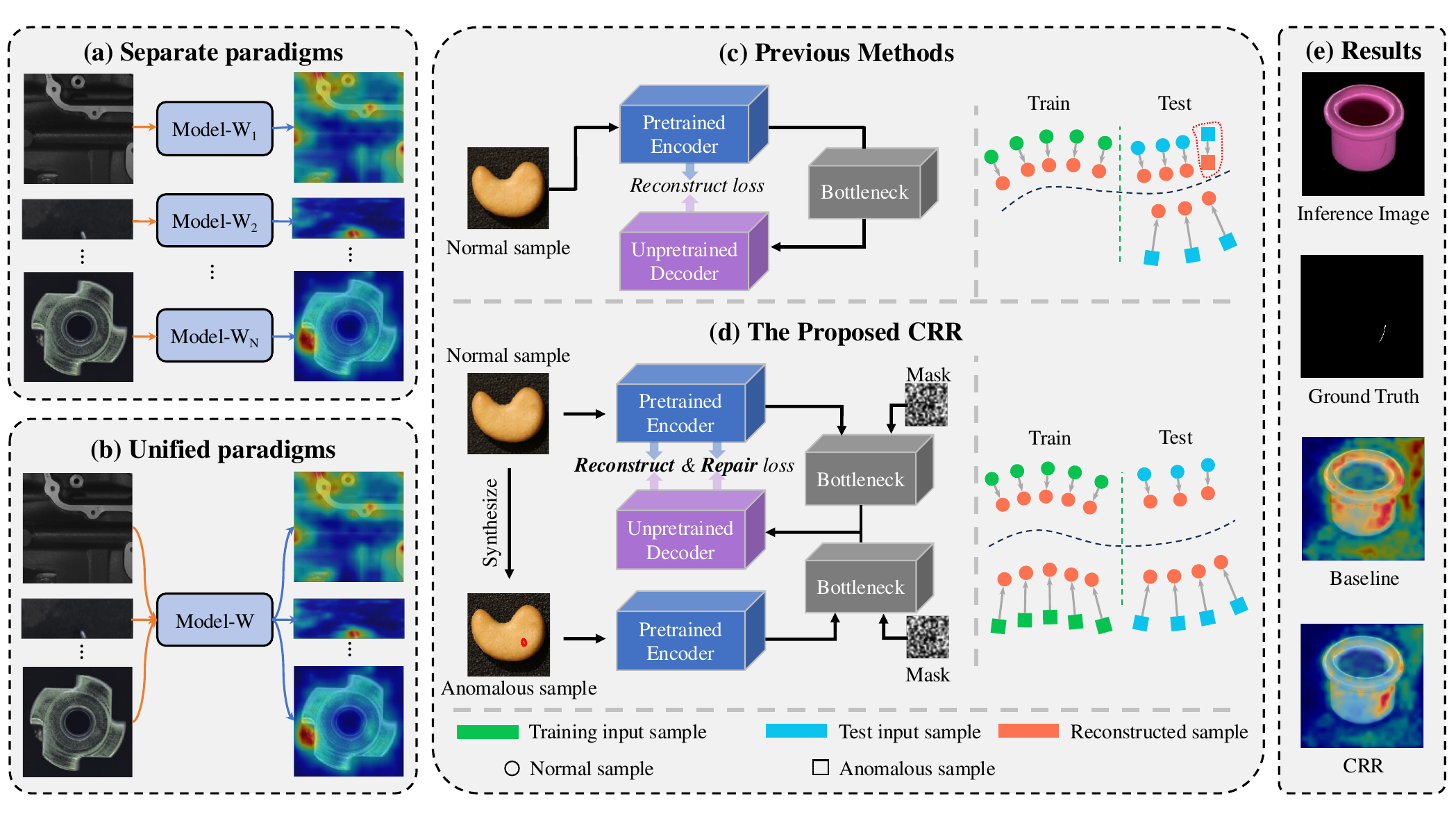}}
\caption{Left: Task setting of IAD and MIAD. (a) The one-class-one-model paradigm assigns distinct weights to each individual category. (b) In contrast, the unified framework employs a single set of shared weights to handle detection tasks across multiple classes. Middle: Comparison between previous methods and the proposed CRR. (c) Previous methods focus only on minimizing the discrepancies between the feature representations from the encoder and decoder on normal samples, which inevitably leads to reduced discrepancies for certain anomalous samples (e.g., anomalous samples within the red dashed box), ultimately causing the identity mapping problem. (d) In contrast, the proposed collaborative reconstruction and repair framework mitigates the identity mapping issue with the assistance of synthesized anomalies in the MIAD task. Right: (e) Comparison of results.}
\label{fig_motivation}
\end{figure}

Industrial anomaly detection (IAD) aims to detect unusual or unexpected patterns in product images that significantly deviate from normative standards. This approach helps reduce manual inspection costs while enhancing product quality inspection efficiency~\cite{zhang2023automated,WOS:001446505200004}, thereby addressing the detection needs of industries such as pollution emissions~\cite{peng2023association}, steelmaking~\cite{WQS_RSA} and photovoltaic manufacturing. As industrial processes continue to improve, collecting sufficient abnormal samples for training becomes increasingly difficult. The types of defects that may occur are unpredictable and diverse. Therefore, industrial anomaly detection often only uses normal samples to train the model.

Traditional approaches to industrial anomaly detection (IAD) build a separate model for each object category, as shown in Fig.~\ref{fig_motivation}(a). However, as products undergo updated and replaced, this one-class-one-model setting entails substantial storage and deployment costs, significantly reducing training efficiency. Recently, UniAD~\cite{uniad} and subsequent studies~\cite{yao2024prior} have proposed training a unified model for multiclass industrial anomaly detection (MIAD), as shown in Fig. \ref{fig_motivation}(b). Under this setting, developing a model to capture the distribution of multi-class objects is fairly challenging.

The current mainstream MIAD algorithms to learning the normal data distribution can be broadly classified into three categories: Augmentation-based~\cite{zavrtanik2021draem,schluter2022nsa,liu2023simplenet}, Reconstruction-based~\cite{tao2022unsupervised,fan2024revitalizing,zhang2024realnet}, and knowledge distillation-based~\cite{zhang2023destseg, guo2024dinomaly,tong2024enhanced,wu2024aekd} methods.
A widely used reconstruction-based scheme assumes that when the decoder is trained to mimic the feature representations from encoder using only normal samples, it will generate feature representations different from those of the encoder network on anomalous samples, as shown in Fig.~\ref{fig_motivation}(c). However, due to the strong generalization capability of the decoder, even with anomalies, the decoder is likely to produce feature representations similar to those of the pretrained encoder.
This phenomenon, known as identity mapping, is illustrated by the anomalous sample highlighted in the red dashed box in Fig.~\ref{fig_motivation}(c).
As a result, such minor discrepancies between the feature representations produced by the encoder and decoder may make detecting anomalies increasingly difficult.
Moreover, in a unified training setting where normal data distribution becomes more intricate, this challenge is further amplified, as illustrated by the baseline result presented in Fig.~\ref{fig_motivation}(e).

Drawing inspiration from DRAEM~\cite{zavrtanik2021draem} and DeSTSeg~\cite{zhang2023destseg}, we propose a method called Collaborative Reconstruction and Repair (CRR), which enables the unpretrained decoder to consistently generate stable normal features, regardless of whether the inputs contain defects, as illustrated in Fig.~\ref{fig_motivation}(d). In this way, the encoder and decoder are reinforced to generate distinct features for anomalous inputs and consistent features for normal inputs.
This method consists of a pretrained encoder, a bottleneck, an unpretrained decoder, and an upsampling segmentation network. 
First, synthetic anomalies are employed as model input to train the decoder, enabling it to generate feature representations consistent with those generated by the encoder for normal images under the same context. Furthermore, a reconstruction constraint is imposed on normal samples to further reduce potential discrepancies between the feature representations produced by the encoder and the decoder.
Second, considering the local and subtle nature of industrial defects, we mask random pixels of the features from encoder, aiming to make decoder infer the missing information based on the neighbor pixels.
Finally, we incorporate a segmentation network to fuse multi-level feature discrepancies, thereby minimizing detection errors resulting from the inherent discrepancies between feature representations from the encoder and decoder, while refining anomalous areas. Fig.~\ref{fig_motivation}(e) also shows that CRR achieves significantly better results than its strong baseline Dinomaly~\cite{guo2024dinomaly}.
The main contributions of this paper are summarized as follows:
\begin{enumerate}
\item We employ normal-sample-based reconstruction and synthesized-anomaly-based repair to generate stable representations of normal data from decoder, thereby producing reliable and precise anomaly localization.
\item We implement feature-level random masking to facilitate the restoration or repair of fine-grained feature representations and utilize a segmentation network to fuse discrepancies across multiple feature levels.
\item We conduct extensive experiments on three popular anomaly detection benchmarks: MVTec-AD, VisA, and Real-IAD. The comprehensive results on these benchmarks across seven metrics demonstrate state-of-the-art performance, thereby substantiating the effectiveness and generalizability of the proposed method. Additionally, we validated its effectiveness on a real-world industrial defect dataset, HSS-IAD.
\end{enumerate}

%% file: sec/2_related_work.tex
\section{RELATED WORK}
\label{related_work}

\subsection{Unsupervised Anomaly Detection}
Recently, significantly superior unsupervised anomaly detectors have been developed. These approaches can be categorized into three mainstream types.
\subsubsection{Augmentation-based methods} These methods synthesize anomalies by adding discontinuous patches or noise to normal images or normal features. DRAEM~\cite{zavrtanik2021draem} generates slightly out-of-distribution appearances using a Perlin noise generator and texture images. CutPaste~\cite{li2021cutpaste} constructs pseudo-anomalous data by cutting out an image patch and pasting it onto a larger image at a random location. NSA~\cite{schluter2022nsa} applies Poisson image editing to seamlessly merge scaled patches of various sizes from different images, generating synthetic anomalies that mimic natural sub-image irregularities. SimpleNet~\cite{liu2023simplenet} generates counterfeit anomaly features by adding Gaussian noise to the features of normal samples.
Using simulated anomalous images along with their corresponding ground truth masks, studies like DRAEM and NSA localize anomalies using segmentation networks. In our approach, we draw on the idea of DRAEM for both anomaly simulation and segmentation.
\subsubsection{Reconstruction-based methods} These methods~\cite{deng2022rd4ad} hold the insight that anomalous regions cannot be properly reconstructed when the model is trained only on normal images. The discrepancy between the input and the reconstructed images can then be used for anomaly localization. Some methods~\cite{tao2022unsupervised,fan2024revitalizing,zhang2024realnet} utilize generative models, including autoencoders and Generative Adversarial Networks (GANs)~\cite{goodfellow2014gan}, to reconstruct normal data, aiming to preserve image category and pixel-wise structural integrity. However, the main problem of these methods is that the model often generalizes well even to anomalies and reconstructs them sufficiently, thus impairing detection capabilities. 
%
\subsubsection{Knowledge distillation--based methods} These methods~\cite{zhang2023destseg, guo2024dinomaly,tong2024enhanced,wu2024aekd} consist of a frozen pre-trained teacher network and a trainable student network. The student network is trained to replicate the features extracted by the teacher network on normal datasets. On abnormal images, the features extracted by the teacher network may diverge from those of the student network. Consequently, the feature discrepancies between the teacher and student networks can be leveraged to detect anomalies.
RD4AD~\cite{deng2022rd4ad} proposed a “reverse distillation” paradigm in which the student network takes the teacher model’s one-class embedding as input and reconstructs multiscale representations from the teacher model.
MRKD~\cite{jiang2023MRKD} employs image-level masking and feature-level masking to restore normal images. 
DeSTSeg~\cite{zhang2023destseg} introduced a denoising encoder-decoder to match the teacher network's features. 
However, the student network in these methods may overgeneralize, producing abnormal features similar to those of the teacher network.
\subsection{Multi-class Anomaly Detection}
Most current methods utilize a one-class-one-model setting, resulting in increased memory and time consumption, which is unsuitable for practical industrial applications. Recently, facing this challenge, multiclass industrial anomaly detection (MIAD) approaches have attracted significant interest. UniAD~\cite{uniad} first introduces a unified framework to cover multiple categories. DiAD~\cite{he2024diad} proposes a diffusion-based anomaly detection framework, utilizing a latent-space semantic-guided network to reconstruct anomalous regions while preserving the original image's semantic information. ViTAD~\cite{zhang2023ViTAD} explores a plain ViT-based symmetric structure, effectively designed step by step from several perspectives on multi-class anomaly detection. Dinomaly~\cite{guo2024dinomaly} utilizes four simple components, foundation transformers, noisy bottleneck, linear attention, and loose reconstruction, to bridge the performance gap between multi-class settings and class-separated setting models. MambaAD~\cite{he2024mambaad} introduced the mamba decoder to capture both long-range and local information and reduce model parameters and computational complexity. However, since these works only adopted normal-sample-based reconstruction, the issue of identity mapping may still be severe.

Different from previous methods, CRR employs collaborative normal image reconstruction and synthetic anomaly repair to ensure that decoder consistently produces stable normal feature representations, regardless of whether there are defects in the input. 
Meanwhile, feature-level random masking is employed to capture normal fine-grained representations, while a segmentation network is utilized to filter out inherent discrepancies. In this way, the proposed method is able to more effectively address the identity mapping problem and differentiate normal from anomalous samples.

%% file: sec/3_method.tex
\section{METHODOLOGY}
\subsection{Problem Definition}
%
%
%
IAD focuses on classifying images as either normal or abnormal while accurately localizing abnormal areas. Given an IAD dataset that contains $N$ classes $\boldsymbol{C}=\left\{C_1, C_2, \cdots, C_N\right\}$, the MIAD setting covers all classes $\boldsymbol{C}$ in one unified model, $\boldsymbol{C_{\textit {Train }}}=\boldsymbol{C_{\textit {Test }}}=\boldsymbol{C}$. The normal images of all classes are used for training, while both normal and defective images are tested together to evaluate the model's capacity.
\subsection{Model overall structure}
\begin{figure}[t]
    \centering
    \includegraphics[width=\textwidth]{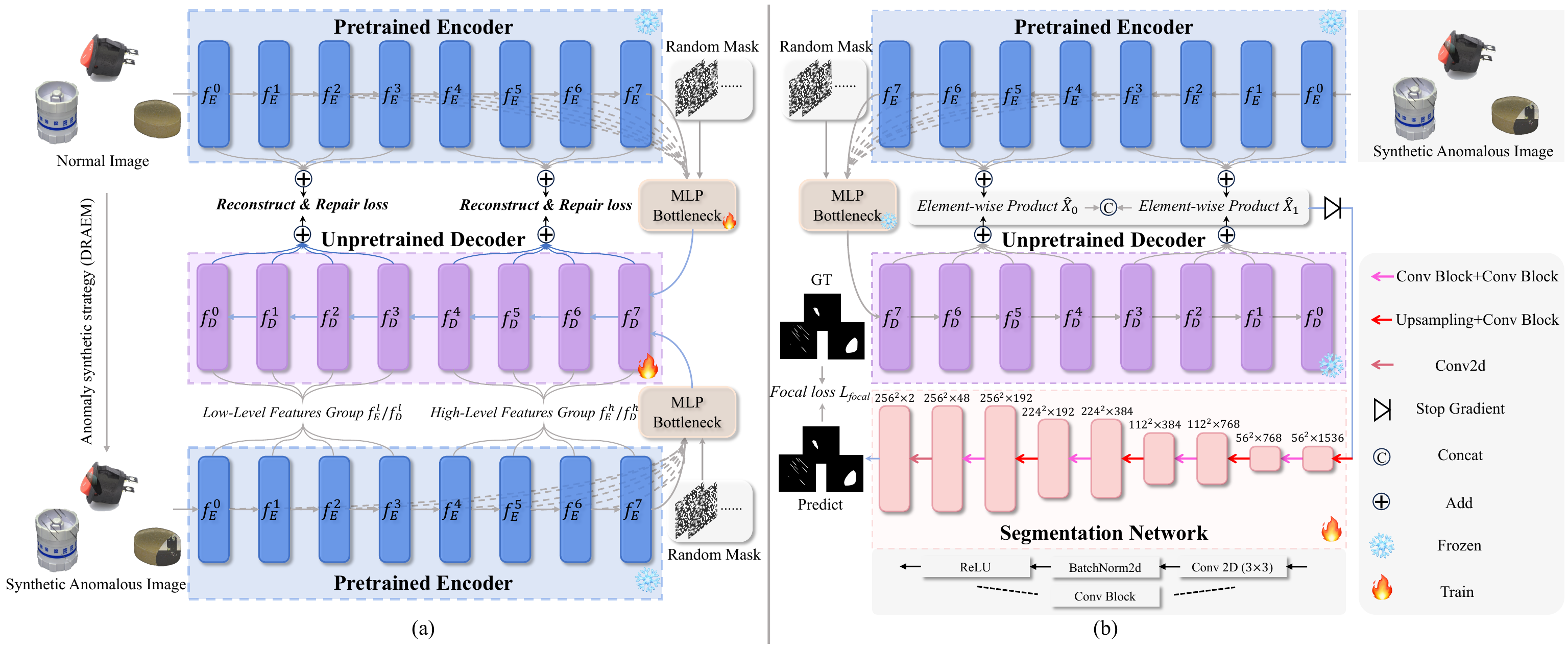}
    \caption{Overview of CRR. In the first step (a), both the bottleneck (MLP) and decoder are trained with normal and synthetic inputs to consistently generate normal features. Some pixel features from the encoder are randomly masked to facilitate the restoration or repair of fine-grained feature representations from visible neighboring patches. In the second step (b), the element-wise product of the encoder's and decoder's normalized outputs is concatenated and used to train the segmentation network. During inference, the anomaly synthesis strategy is not applied to the test images.}
    \label{fig_frame}
\end{figure}
Denoting the normal and abnormal features in the encoder and decoder as $\boldsymbol{f}_{\!E,n}$, $\boldsymbol{f}_{\!D,n}$, $\boldsymbol{f}_{\!E,a}$ and $\boldsymbol{f}_{\!D,a}$, most existing methods based on feature reconstruction aim to minimize the discrepancies between $\boldsymbol{f}_{\!E,n}$ and $\boldsymbol{f}_{\!D,n}$, as formulated below:
\begin{equation}
J = \mathcal{D}\left(\boldsymbol{f}_{\!E,n}, \boldsymbol{f}_{\!D,n}\right),
\label{eq:Dis}
\end{equation}
where $\mathcal{D}(\cdot,\cdot)$ denotes the cosine similarity function that calculates the discrepancy between two sets of features. Previous studies~\cite{deng2022rd4ad,guo2024dinomaly} assumed that the discrepancies between $\boldsymbol{f}_{\!E,a}$ and $\boldsymbol{f}_{\!D,a}$ would remain large and relatively unaffected during the minimization process for $\boldsymbol{f}_{\!E,n}$ and $\boldsymbol{f}_{\!D,n}$. However, due to identity mapping, the similarity between $\boldsymbol{f}_{\!E,a}$ and $\boldsymbol{f}_{\!D,a}$ may increase, leading to high prediction uncertainty.

To address the identity mapping issue, this study proposes the CRR approach, which enhances prediction certainty by reconstructing normal features to normal and repairing abnormal features to normal collaboratively. Consequently, when actual anomalies are input into the model, the resulting discrepancies between $\boldsymbol{f}_{\!E,a}$ and $\boldsymbol{f}_{\!D,a}$ can accurately indicate the location of the anomalies. Since abnormal images cannot be used during the training phase, CRR employs a data augmentation strategy \cite{zavrtanik2021draem} to introduce synthetic anomalies into foreground of normal images, generating synthetic anomalous samples. 

As depicted in Fig. \ref{fig_frame}, the proposed CRR consists of three primary components: Normal Reconstruction and Anomaly Repair (NRAR), Feature Masking (FM), and Segmentation Network (SN). Initially, normal images are fed into a pretrained encoder to extract features, which serve as supervisory signals. The synthetic anomalous samples are then used as input for training an decoder to learn contextual relationships and convert local abnormal features back to normal, while preserving normal areas. Additionally, normal samples are used to train the decoder to reconstruct normal features. Subsequently, some pixel features from encoder are randomly masked to reinforce the decoder to restore or repair grained features from visible neighboring patches. Once this step is completed, the decoder module is fixed. Both the decoder and encoder networks process the synthetic anomaly images to optimize the parameters in the segmentation network, allowing for the localization of anomalous regions. The remainder of this section provides a detailed explanation of NRAR, FM, and SN, followed by an outline of the inference phase, which specifies the procedure for detecting and localizing anomalies.
\subsection{Normal Reconstruction and Anomaly Repair (NRAR)} 
Theoretically, as long as a substantial difference is ensured between the feature representations of the encoder and decoder in the anomalous regions of synthetic anomalous samples, the issue of overgeneralization can be mitigated.
However, our preliminary experimental results indicate that repairing anomalous features to resemble normal features results in improved performance. One possible explanation is that feature representations from the encoder on synthetic anomalous samples varies due to the different positions of anomalous regions within the normal image. 
Furthermore, using these variable features as input to the decoder produces more diverse feature representations, which makes it increasingly difficult to keep them distinct.
In contrast, utilizing normal feature representations distilled by the encoder as a constant supervisory signal enables the decoder to repair the local anomalies as normal.
As the encoder has been pre-trained on a large dataset, it can generate discriminative feature representations in both normal and anomalous regions. Therefore, the decoder will generate different feature representations from those by the encoder during inference. We also optimize the decoder to reconstruct the normal features.
Moreover, reconstruction of normal samples and repair of synthetic anomalies encourages the decoder to learn both low-scale information (i.e., texture and edge) and large-scale information (i.e., structure and orientation) of normal samples in detail.

Following prior work~\cite{guo2024dinomaly}, the encoder $E$ is a standard pre-trained ViT-Base/14 network with 12 Transformer layers, extracting feature maps from the eight middle-level layers, denoted as $\boldsymbol{f}_{\!E}^{i} \left ( i=0\sim 7 \right )$ with size $h_0 \times w_0$. $E$ is parameterized by $\theta _{E}$ that is frozen during the training stage and i represents the i-th block in $E$. The bottleneck $B$ is a simple MLP (a.k.a. feed-forward network, FFN) that integrates the encoders' representations, utilizing randomly initialized weights $\theta_B$. The decoder $D$, like the encoder, includes 8 unpretrained Transformer layers and is randomly initialized by parameter $\theta _{D}$. The corresponding feature representation is denoted as $\boldsymbol{f}_{\!D}^{i} \left ( i=0\sim 7 \right )$.
Consistent with the earlier study~\cite{guo2024dinomaly}, we group the features into low-semantic-level and high-semantic-level groups through Eq. \eqref{eq:fE_lh} and \eqref{eq:fD_lh}. Specifically, $\textit{vec}\left( \cdot  \right )$ denotes flatten operation. Thus, the feature discrepancy between the encoder and decoder is quantified by the average cosine similarity between the two groups, as shown in Eq. \eqref{eq:d_fE_fD}. In particular, $k=l$ or $h$. To collaboratively reconstruct normal features $\boldsymbol{f}_{\!D,n}$ and repair abnormal features $\boldsymbol{f}_{\!D,a}$ to the normal data manifold, we align them with $\boldsymbol{f}_{\!E,n}$. The loss function is defined as the sum of the feature discrepancies for both normal and abnormal samples, as shown in Eq. \eqref{eq:loss1}.
%
%
\begin{equation}
\begin{aligned}
\boldsymbol{f}_{\!E}^{l} &= \frac{1}{4}\sum_{i=0}^{3}\boldsymbol{f}_{\!E}^{i}, &
\boldsymbol{f}_{\!E}^{h} &= \frac{1}{4}\sum_{i=4}^{7}\boldsymbol{f}_{\!E}^{i},
\end{aligned}
\label{eq:fE_lh}
\end{equation}
\begin{equation}
\begin{aligned}
\boldsymbol{f}_{\!D}^{l} &= \frac{1}{4}\sum_{i=0}^{3}\boldsymbol{f}_{\!D}^{i}, &
\boldsymbol{f}_{\!D}^{h} &= \frac{1}{4}\sum_{i=4}^{7}\boldsymbol{f}_{\!D}^{i},
\end{aligned}
\label{eq:fD_lh}
\end{equation}
\begin{equation}
\mathcal{D}\left ( \boldsymbol{f}_{\!E}, \boldsymbol{f}_{\!D} \right ) = \sum_{k\in \left \{ l,h\right \} } \left( 1 - \frac{\textit{vec} \left ( \boldsymbol{f}_{\!E}^{k} \right ) \cdot \textit{vec} \left ( \boldsymbol{f}_{\!D}^{k} \right )}{\left\|\textit{vec} \left ( \boldsymbol{f}_{\!E}^{k} \right )\right\|\left\|\textit{vec} \left ( \boldsymbol{f}_{\!D}^{k} \right )\right\|} \right),
\label{eq:d_fE_fD}
\end{equation}
\begin{equation}
L_{\textit{cos}} = \mathcal{D}\left ( \boldsymbol{f}_{\!E,n}, \boldsymbol{f}_{\!D,n} \right ) + \mathcal{D}\left ( \boldsymbol{f}_{\!E,n}, \boldsymbol{f}_{\!D,a} \right ).
\label{eq:loss1}
\end{equation}
\subsection{Feature Masking (FM)}
The detection of subtle defects in real-world scenarios presents significant challenges, as these defects only induce local alterations in the image's contextual information. To mitigate the propagation of anomalous perturbations to the decoder and enable accurate perception of fine-grained features, we implement a strategy of randomly masking all areas of the feature of encoder. This strategy leverages local information to refine the restored features, ensuring the preservation of feature details and generating "normal-like" features, thereby enhancing the representation power of local image information. Furthermore, this approach accentuates the imbalance between input and supervisory signals, consequently alleviating the issue of identity mapping.

Feature masking is implemented using Masked Generative Distillation (MGD)~\cite{yang2022MGD}, which randomly masks all areas of $\boldsymbol{f}_{\!E}^{7}$, regardless of whether they are abnormal or normal (see Fig. \ref{fig_frame}). 
Subsequently, the bottleneck $B$ and decoder $D$ is employed to restore the masked features, generating the full normal features $\boldsymbol{f}_{\!D}^i$. The process can be formulated as follows:
\begin{align}
    M(h, w) &= \begin{cases}
        0, & \text{if } R(h, w)<\lambda \\ 
        1, & \text{otherwise}
    \end{cases} \label{eq:Mask} \\
    \boldsymbol{f}_{\!D}^i &= D\left (B \left(\boldsymbol{f}_{\!E}^{7} \odot M,\theta_B\right), \theta_D\right),
    \label{eq:fD_i_m}
\end{align}
where $R(h, w)$ denotes a random value within the range $\left ( 0,1 \right ) $ at the image coordinates $\left ( h,w \right ) $, and $M$ denotes the generated mask. 
The parameter $\lambda$, which represents the masking ratio, is determined through the ablation study presented in Section~\ref{sec_ablation_mask_rates}, while $\odot$ denotes element-wise multiplication.
\subsection{Segmentation Network (SN)}
\label{SN-}
In previous study~\cite{guo2024dinomaly}, the anomaly score for each pixel is derived by directly summing the cosine distances from two groups of features. However, the performance could be improved when there are inherent discrepancies between feature representations from encoder and decoder. To address these issues, we appended an upsampling segmentation network to filter out the aforementioned discrepancies and refine the predicted regions.
\begin{algorithm}[t]
  \caption{Collaborative Reconstruction and Repair.}
  \textbf{\# Training Stage}\\
  \textbf{Input:}
  Training dataset $\boldsymbol{C_{\textit {Train }}}$, encoder network $E$, bottleneck module $B$, decoder network $D$, segmentation network $S$, hyperparameter $\lambda$, and their parameters $\left \{ \theta _E,\theta _B,\theta _D,\theta _S \right \} $\\
  \textbf{Output:} 
  \makebox[3cm][l]{The parameters $\left\{\theta _B, \theta _D, \theta _S \right\}$}
  \begin{algorithmic}[1]
    \STATE Initialize $\theta _E$ with pretrained weights\\
    Initialize $\left \{\theta _B,\theta _D,\theta _S \right \} $ with random weights
    \FOR{$iter\_num=1$ \textbf{to} $n\_iters$}
      \STATE Randomly sample a batch of normal samples $I_n$
      \STATE Generate synthetic abnormal samples $I_a$
      \STATE Generate low-semantic-level and high-semantic-level grouped features of $I_n$ and $I_a$, $\boldsymbol{f}_{\!E,n}^{l}$, $\boldsymbol{f}_{\!E,n}^{h}$, $\boldsymbol{f}_{\!D,n}^{l}$, $\boldsymbol{f}_{\!D,n}^{h}$, $\boldsymbol{f}_{\!D,a}^{l}$, $\boldsymbol{f}_{\!D,a}^{h}$ according to Equation~\eqref{eq:fE_lh}, \eqref{eq:Mask}, \eqref{eq:fD_i_m}, \eqref{eq:fD_lh}
      \STATE Calculate the feature discrepancies for reconstructing normal samples and repairing abnormal samples, denoted as  $\mathcal{D}\left(\boldsymbol{f}_{\!E,n}, \boldsymbol{f}_{\!D,n}\right)$ and $\mathcal{D}\left(\boldsymbol{f}_{\!E,n}, \boldsymbol{f}_{\!D,a}\right)$, using Equation~\eqref{eq:d_fE_fD}
      
      \STATE Calculate the total loss $L$ using Equation~\eqref{eq:loss1}
      \STATE Update $\left \{\theta _B,\theta _D \right \} $ iteratively using gradient step
    \ENDFOR
    \FOR{$iter\_num=1$ \textbf{to} $n\_iters'$}
      \STATE Repeat steps 3-5 from the first training stage
      \STATE Calculate feature $\hat{X}$ according to $\left(\boldsymbol{f}_{\!E}^{l}, \boldsymbol{f}_{\!D}^{l}\right)$ and ${\left(\boldsymbol{f}_{\!E}^{h}, \boldsymbol{f}_{\!D}^{h} \right)}$
      \STATE Calculate the predicted value $Y$ of the segmentation network $S$
      \STATE Calculate the focal loss $L_{focal}$ according to Equation~\eqref{eq:Loss_focal}
      \STATE Perform a gradient descent step to update $\left \{\theta _S \right \}$
    \ENDFOR
  \end{algorithmic}
  \textbf{\# Inference Stage}\\
  \textbf{Input:} 
  Testing dataset $\boldsymbol{C_{\textit {Test }}}$, the parameters $\left \{\theta _B, \theta _D, \theta _S \right \}$ with their saved weights\\
  \textbf{Output:} \makebox[3cm][l]{Anomaly scores $S_{AL}$ and $S_{AD}$}
  \begin{algorithmic}[1]
    \STATE Repeat steps 3, 5, and 6 from the first training stage to calculate the feature discrepancy $\mathcal{D}\left(\boldsymbol{f}_{\!E}, \boldsymbol{f}_{\!D}\right)$
    \STATE Generate an anomaly detection mask $S_{\mathrm{seg}}$
    \STATE Calculate the anomaly scores $S_{AL}\left ( h,w \right )$ and $S_{AD}$ using Equation~\eqref{eq:S_AL} and \eqref{eq:S_AD}
  \end{algorithmic}
\label{algorithm}
\end{algorithm}

To mitigate the risk of gradient explosion during the optimization of the segmentation network, we froze the weights of both the encoder and decoder. The synthetic anomalous image serves as input for the encoder, with the corresponding binary anomaly mask serving as the ground truth. The similarities of the feature maps $\left(\boldsymbol{f}_{\!E}^{l}, \boldsymbol{f}_{\!D}^{l}\right)$ and ${\left ( \boldsymbol{f}_{\!E}^{h}, \boldsymbol{f}_{\!D}^{h}   \right )}$ are computed through element-wise multiplication, resulting in $\hat X_{0}$ and $\hat X_{1}$, which are subsequently concatenated to form $\hat{X}$. This feature $\hat{X}$ is then fed into the segmentation network. The segmentation network $S$, initialized with random weights $\theta_S$, consists of four convolutional blocks and four upsampling modules, with its output size matching that of the ground truth (see Fig. \ref{fig_frame}). Compared to non-parametric upsampling methods, the segmentation network enables adaptive feature fusion, thereby enhancing the precision of localized regions.

Given the issue of area imbalance between normal and abnormal regions in images, we implemented the focal loss~\cite{ross2017focalloss} to optimize the segmentation network, thereby enhancing the model's ability to concentrate on the segmentation of challenging samples. Specifically, we minimized the focal loss between the ground truth $G$ of the synthetic image and the predicted value $Y$ of the model, as expressed as follow:
\begin{equation}
    L_{focal}=-\alpha_t\left(1-p_t\right)^\gamma \log \left(p_t\right),
    \label{eq:Loss_focal}
\end{equation}
where $p_t$ denotes the predicted probability for pixel category. It equals the predicted probability $p$ when the actual label of the corresponding pixel in $G$ is 1. Conversely, when the actual label is 0, $p_t$ is calculated as $1-p$. Additionally, the hyperparameters $\alpha_t$ and $\gamma$ are employed to modulate the degree of weighting. In conclusion, our optimization goal is to assign higher weights to subtle abnormal regions over normal ones in the loss function, thereby improving the accuracy of abnormal segmentation.

\subsection{Inference}
After optimizing the decoder with the proposed strategy, the decoder module is endowed with the capability to consistently output normal feature representations from local to global scales, regardless of the presence of anomalies in the image. During the inference stage, the test image is fed into the encoder. The discrepancies between features $\boldsymbol{f}_{\!E}$ and $\boldsymbol{f}_{\!D}$, denoted as $\mathcal{D}\left ( \boldsymbol{f}_{\!E}, \boldsymbol{f}_{\!D} \right )$, can provide compelling evidence for localizing anomalies, as shown in Eq.~\eqref{eq:d_fE_fD}. However, we have observed that the predicted regions can be further refined for greater precision using a segmentation network.
The similarity map $\hat{X}$ is fed into the segmentation network to generate an anomaly score map of size $h \times w$ (i.e., $1/14$ of $h_0$ and $w_0$), denoted as $S_{\mathrm{seg}}(h, w)$.
To preserve the precise distribution in $\mathcal{D}\left ( \boldsymbol{f}_{\!E}, \boldsymbol{f}_{\!D} \right )$ and the accurate localization in $S_{\mathrm{seg}}\left ( h,w \right ) $, we sum $\mathcal{D}\left ( \boldsymbol{f}_{\!E}, \boldsymbol{f}_{\!D} \right )$ and $S_{\mathrm{seg}}\left ( h,w \right ) $, weighted by the hyperparameters $\lambda_1$ and $\lambda_2$, and upsampled to the input size to produce the final anomaly score map:
\begin{equation}
    S_{AL}\left ( h,w \right ) =\mathit{\Phi} (\lambda_1 \cdot \mathcal{D}\left ( \boldsymbol{f}_{\!E},\boldsymbol{f}_{\!D} \right ) +\lambda_2 \cdot S_{\mathrm{seg}}\left ( h,w \right ) ),
\label{eq:S_AL}
\end{equation}
where $\mathit{\Phi}$ function performs a bilinear up-sampling operation.

The image-level anomaly score is derived by averaging the highest $T$ values from the anomaly score map $S_{AL}$, where $T$ is a configurable hyperparameter. Hence, $S_{AD}$ is achieved by:
\begin{equation}
    S_{AD} = \frac{1}{T} \sum_{i=1}^{T} S_{AL}(h, w).
\label{eq:S_AD}
\end{equation}

Notably, feature-level masking strategies are applied during training. In the testing stage, it is also crucial to perform feature masking to ensure that the test samples align with the same domain as the training samples. A comprehensive overview of the proposed method is presented in Algorithm~\ref{algorithm}.

%% file: sec/4_experiments.tex
\section{RESULTS AND DISCUSSION}
\subsection{Experimental Settings}
In this section, a series of experiments are conducted on the MVTec-AD~\cite{bergmann2019mvtec}, VisA~\cite{zou2022visa}, and Real-IAD~\cite{wang2024real} datasets to evaluate the performance of CRR and demonstrate the role of its individual components.
Additionally, CRR is evaluated on the HSS-IAD~\cite{wang2025hss} dataset to validate its effectiveness in real-world industrial scenarios.
\subsubsection{Datasets Descriptions}
\textbf{MVTec-AD} is a widely used dataset for MIAD. 
The dataset consists of 3,629 normal images for training and a test set of 1,725 images, of which 467 are normal and 1,258 are anomalous.
\textbf{VisA} features 12 different object categories.
It contains 8,659 normal images for training and 2,162 images for evaluation, including 962 normal and 1,200 anomalous images in the test set.
\textbf{Real-IAD} covers 30 distinct object categories, with a training set consisting of 36,465 normal images and a test set comprising 63,256 normal and 51,329 abnormal samples, following the official data split.
\textbf{HSS-IAD} (Heterogeneous Same-Sort Industrial Anomaly Detection) dataset is a real-world benchmark for industrial anomaly detection. It consists of various same-sort components commonly found in manufacturing, including electrical commutators, magnetic tiles, flat sheet steel, and engine castings. The training set comprises 9,385 normal samples, while the test set contains 2,017 normal and 1,831 anomalous samples.

\subsubsection{Metrics}
Following prior works~\cite{he2024mambaad}, we adopt eight evaluation metrics. For anomaly detection and segmentation, we report the Area Under the Receiver Operating Characteristic Curve (AU-ROC), Average Precision (AP)~\cite{zavrtanik2021draem}, and F1-score-max ($F_{1}$-max)~\cite{zou2022visa}.
Additionally, we report the Area Under the Per-Region-Overlap (AU-PRO) curve to evaluate segmentation performance.
We further calculate the mean value of these seven evaluation metrics (denoted as mAD) to represent the model's comprehensive capability~\cite{zhang2023ViTAD}. The results for a dataset are averaged across all classes.

\begin{table}[t]
    \centering
    \setlength{\tabcolsep}{1.5mm}    
    \caption{Quantitative Results on different AD datasets for multi-class setting.}
    \resizebox{\linewidth}{!}{
        \begin{tabular}{ccccccccccc}
            \toprule
            \multirow{2}[2]{*}{Dateset} & \multirow{2}[2]{*}{Method} & \multirow{2}[2]{*}{Public} & \multicolumn{3}{c}{Image-level} & \multicolumn{4}{c}{Pixel-level}& \multirow{2}[2]{*}{\textbf{mAD}} \\
            \cmidrule(r){4-6} \cmidrule(l){7-10} 
            &&&\multicolumn{1}{c}{AUROC} & \multicolumn{1}{c}{AP} & \multicolumn{1}{c}{$F_1$-max} & \multicolumn{1}{c}{AUROC} & \multicolumn{1}{c}{AP} & \multicolumn{1}{c}{$F_1$-max} & \multicolumn{1}{c}{AUPRO} \\
                \hline
            
            \multirow{6}[0]{*}{MVTec-AD~\cite{bergmann2019mvtec}} & RD4AD~\cite{deng2022rd4ad}  & CVPR'22   & 94.6  & 96.5  & 95.2  & 96.1  & 48.6  & 53.8  & 91.1 &82.3\\
            & UniAD~\cite{uniad} & NeurIPS'22    & 96.5  & 98.8  & 96.2  & 96.8  & 43.4  & 49.5  & 90.7& 81.7\\
            & SimpleNet~\cite{liu2023simplenet} & CVPR'23   & 95.3  & 98.4  & 95.8  & 96.9  & 45.9  & 49.7  & 86.5& 81.2\\            
            & DiAD~\cite{he2024diad}    & AAAI'24   & 97.2  & 99.0  & 96.5  & 96.8  & 52.6  & 55.5  & 90.7& 84.0    \\
            & MambaAD~\cite{he2024mambaad}  & NeurIPS'24    & 98.6  & 99.6  & 97.8  & 97.7  & 56.3  & 59.2  & 93.1  & 86.0  \\            
            & Dinomaly~\cite{guo2024dinomaly}   & CVPR'25   & \underline{99.6} & \underline{99.8} & \underline{99.0} & \textbf{98.4} & \underline{69.3} & \textbf{69.2} & \underline{94.8} & \underline{90.0}\\
            & \textbf{CRR} (Ours)   & - & \textbf{99.7} & \textbf{99.9} & \textbf{99.2} & \textbf{98.4} & \textbf{71.4} & \underline{68.9} & \textbf{95.5}& \textbf{90.4}\\
            \hline
            
            \multirow{6}[0]{*}{VisA~\cite{zou2022visa}} & RD4AD~\cite{deng2022rd4ad}    & CVPR'22   & 92.4  & 92.4  & 89.6 & 98.1 & 38.0  & 42.6  & 91.8 & 77.8 \\
            & UniAD~\cite{uniad}    & NeurIPS'22    & 88.8  & 90.8  & 85.8  & 98.3  & 33.7  & 39.0  & 85.5& 74.6    \\            
            & SimpleNet~\cite{liu2023simplenet} & CVPR'23   & 87.2  & 87.0  & 81.8  & 96.8  & 34.7  & 37.8  & 81.4  &72.4 \\
            & DiAD~\cite{he2024diad}    & AAAI'24  & 86.8  & 88.3  & 85.1  & 96.0  & 26.1  & 33.0  & 75.2& 70.1\\
            & MambaAD~\cite{he2024mambaad}  & NeurIPS'24    & 94.3  & 94.5  & 89.4  & 98.5  & 39.4  & 44.0  & 91.0  & 78.7  \\
            & Dinomaly~\cite{guo2024dinomaly}   & CVPR'25   & \underline{98.7} & \underline{98.9} & \underline{96.2} & \underline{98.7} & \underline{53.2} & \underline{55.7} & \underline{94.5} & \underline{85.1}\\
            & \textbf{CRR} (Ours)   & - & \textbf{99.2} & \textbf{99.3} & \textbf{97.0}  & \textbf{98.8} & \textbf{55.6}  & \textbf{57.0} & \textbf{96.3} &\textbf{86.2}\\
            \hline
            
            \multirow{6}[0]{*}{Real-IAD~\cite{wang2024real}}    & RD4AD~\cite{deng2022rd4ad}    & CVPR'22   & 82.4  & 79.0  & 73.9  & 97.3  & 25.0  & 32.7  & 89.6 & 68.6 \\
            & UniAD~\cite{uniad}    & NeurIPS'22    & 83.0  & 80.9  & 74.3  & 97.3  & 21.1  & 29.2  & 86.7 &67.5    \\
            & SimpleNet~\cite{liu2023simplenet} & CVPR'23   &57.2  & 53.4  & 61.5  & 75.7  & \;\;2.8   & \;\;6.5   & 39.0 & 42.3 \\
            & DiAD~\cite{he2024diad}    & AAAI'24   &75.6  & 66.4  & 69.9  & 88.0  &  \;\;2.9   &  \;\;7.1   & 58.1 & 52.6   \\
            & MambaAD~\cite{he2024mambaad}  & NeuIPS'24 & 86.3  & 84.6  & 77.0  & 98.5  & 33.0  & 38.7  & 90.5  & 72.7  \\
            & Dinomaly~\cite{guo2024dinomaly}   & CVPR'25   & \underline{89.3} & \underline{86.8} & \underline{80.2} & \underline{98.8} & \underline{42.8} & \underline{47.1} & \underline{93.9} & \underline{77.0}\\   
            & \textbf{CRR} (Ours)   & - &\textbf{91.3}  & \textbf{89.7} & \textbf{82.6} & \textbf{99.2} & \textbf{54.0}  & \textbf{54.2}  & \textbf{95.8} &\textbf{81.0}\\
            \hline
            
            \multirow{6}[0]{*}{HSS-IAD~\cite{wang2025hss}}  
            & DRAEM\cite{zavrtanik2021draem}    & ICCV'21   & 63.7  & 57.5  & 74.1  & 70.5  & \pzo 8.2  & 11.4  & 24.1  & 44.2  \\ 
            & RD4AD~\cite{deng2022rd4ad}  & CVPR'22 & 69.2  & 74.2  & 77.3  & 79.9  & 16.3  & 20.8 & \underline{58.1} & 56.5   \\
            & UniAD~\cite{uniad} & NeurIPS'22 & 63.4  & 71.5  & 75.0  & 80.7  & 13.4  & 17.2  & 49.6  & 53.0  \\
            & SimpleNet~\cite{liu2023simplenet}  & CVPR'23   & 54.3  & 62.2  & 71.4  & 54.2  & \pzo9.5  & 12.4  & 20.5  & 40.6  \\            
            & DeSTSeg~\cite{zhang2023destseg}    & CVPR'23   & 73.6  & 77.8  & 79.0  & \underline{84.0}  & 19.8  & 23.5  & 55.6  & 59.0  \\
            & Dinomaly~\cite{guo2024dinomaly}   & CVPR'25  & \underline{77.7} & \underline{79.9} & \underline{81.1} & 83.8 & \underline{22.5} & \underline{25.7} & 54.8 & \underline{60.8}\\
            & \textbf{CRR} (Ours)   & - & \textbf{80.4} & \textbf{83.3} & \textbf{81.2} & \textbf{88.8} & \textbf{24.2} & \textbf{29.1} & \textbf{64.9}& \textbf{64.6}  \\
            \bottomrule
            
        \end{tabular}%
    }
    \label{tab:results}%
\end{table}

\subsubsection{Implementation Details}
 The ViT-Base/14 model (patch size 14), pre-trained using DINOv2-R \cite{oquab2023dinov2}, serves as the default encoder. The Bottleneck's drop rate is initially set at 0.4 and is reduced to 0.2 for the HSS-IAD dataset (to preserve more feature information for industrial parts with poor semantic integrity under complex conditions). Input images are resized to $448^2$ and center-cropped to $392^2$, ensuring the feature map ($28^2$) is sufficiently large for anomaly localization. The StableAdamW optimizer \cite{wortsman2023StableAdamW}, incorporating AMSGrad,
 is employed with a learning rate ($lr$) of 2e-3, $\beta$ values of (0.9, 0.999), a batch size of 8, and weight decay ($wd$) of 1e-4 during the first stage. During the second stage, the AdamW optimizer is used with a learning rate of 1e-4 and a batch size of 16.
The encoder-decoder network undergoes training for 30,000 iterations on HSS-IAD, 10,000 iterations on MVTec-AD and VisA, and 50,000 iterations on Real-IAD during the first stage. For the segmentation network, training is conducted for 5,000 iterations on HSS-IAD, 10,000 iterations on MVTec-AD, 4,000 iterations on VisA, and 8,000 iterations on Real-IAD.
Empirically, $\lambda_1$ is 0.7 and $\lambda_2$ is 0.3.
\subsection{Comparison with SoTAs on Different AD datasets}
We compare the proposed CRR with several state-of-the-art (SoTA) methods on a range of datasets utilizing both image-level and pixel-level metrics. Notably, UniAD~\cite{uniad}, which first introduced this practical setting, along with DiAD~\cite{he2024diad} based on diffusion reconstruction and Dinomaly~\cite{guo2024dinomaly} relying on feature reconstruction, are all designed for MIAD tasks. Meanwhile, RD4AD~\cite{deng2022rd4ad}, which also utilizes feature reconstruction, and SimpleNet~\cite{liu2023simplenet}, leveraging feature-level pseudo-anomalies, are tailored to traditional class-separated IAD scenarios. To ensure fair evaluation, we extend the official codes of these methods for unified training under the MIAD setting.
\subsubsection{Quantitative Comparisons with SoTAs}
Experimental results are presented in Table \ref{tab:results}, where CRR outperforms the compared methods by a significant margin across almost all datasets and metrics.
%
On the widely used MVTec-AD dataset, CRR achieves a new SoTA with image-level performance of \textbf{99.7}/\textbf{99.9}/\textbf{99.2} and pixel-level performance (AP/AUPRO) of \textbf{71.4}/\textbf{95.5}, representing improvements of \textbf{0.1}/\textbf{0.1}/\textbf{0.2} and \textbf{2.1}/\textbf{0.7}, respectively, compared to the strong baseline model Dinomaly.
Additionally, we achieve a \textbf{0.4} increase compared to the advanced Dinomaly on the mAD metric. Additionally, we have discovered that models designed for class-separate IAD tasks did not achieve superior performance in multi-class scenarios, as evidenced by comparisons with SimpleNet and other models. This may be attributed to model overfitting or single-class-specific training strategies. 

On the challenging VisA dataset, CRR achieves image-level performance of \textbf{99.2}/\textbf{99.3}/\textbf{97.0} and pixel-level performance of \textbf{98.8}/\textbf{55.6}/\textbf{57.0}/\textbf{96.3}, demonstrating improvements of \textbf{0.5}/\textbf{0.4}/\textbf{0.8} and \textbf{0.1}/\textbf{2.4}/\textbf{1.3}/\textbf{1.8}, respectively. Notably, our CRR achieves an improvement of \textbf{1.1} in the overall mAD metric compared to the previous SoTA.
On the Real-IAD dataset,
we achieve image-level performance of \textbf{91.3}/\textbf{89.7}/\textbf{82.6} and pixel-level performance of \textbf{99.2}/\textbf{54.0}/\textbf{54.2}/\textbf{95.8}, demonstrating improvements of \textbf{2.0}/\textbf{2.9}/\textbf{2.4} and \textbf{0.4}/\textbf{11.2}/\textbf{7.1}/\textbf{1.9}, respectively.
Notably, as indicated by Dinomaly~\cite{guo2024dinomaly}, enlarging the input resolution of comparison methods fails to yield performance gains and even deteriorates their results, especially on image-level metrics.
For the overall mAD metric, our CRR shows a \textbf{4.0} improvement compared to the previous SoTA.
This indicates the generalizability, versatility, and efficacy of our method in extremely complex scenarios. Per-class performances are presented in Appendix~\ref{sec:perclass}.
%
\begin{figure}[t]
	\centering
	\includegraphics[width=\textwidth]{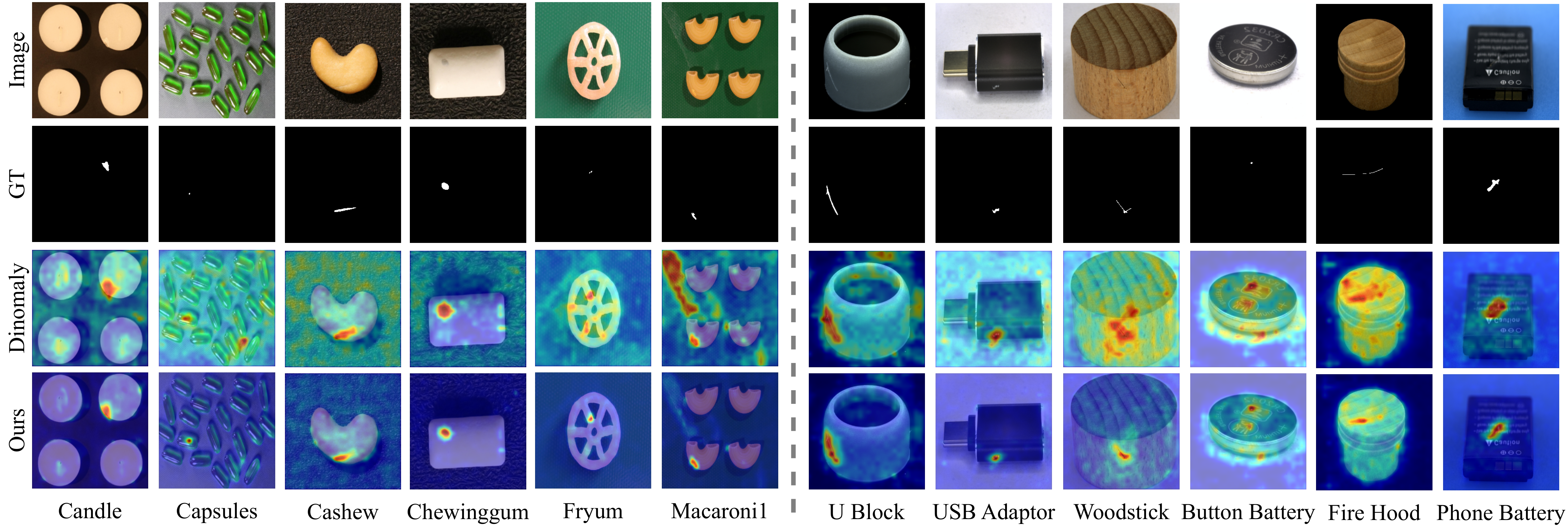}
	\caption{Qualitative visualization for pixel-level anomaly segmentation on VisA and Real-IAD datasets.}
	\label{fig::visual}
\end{figure}
\subsubsection{Qualitative Comparison with SoTAs}
To further assess the accuracy of our proposed approach in anomaly localization, we conducted qualitative evaluations on VisA and Real-IAD datasets. As shown in Fig. \ref{fig::visual}, the left and right sides respectively display visualizations of Dinomaly and CRR across different categories within the VisA and Real-IAD datasets. Compared to the SoTA method (Dinomaly), our CRR method consistently achieves more precise and compact anomaly localization, with reduced edge uncertainty and fewer false anomaly responses in normal regions.
Additional qualitative results for each class are provided in Appendix~\ref{sec:visual}.
%
\subsection{Ablation Study}
\subsubsection{Network architecture}
To verify the effectiveness of the CRR components and evaluate the impact of hyperparameter selection, we conducted comprehensive experiments on Real-IAD under a unified case. Specifically, our three design elements include: employing NRAR to generate stable normal features from decoder, utilizing FM to mitigate the propagation of anomalous perturbations, and appending SN to supplement the feature similarity comparison strategy. We take Dinomaly~\cite{guo2024dinomaly} as the baseline and report the effectiveness of the CRR components in Table \ref{tab:abvisa}. (a) Comparing experiments 1 and 2, it can be found that applying feature-level masking improves performance. Pixel-level performance improved significantly, due to the stronger representation power of local image information after feature-level masking. (b) The comparisons between experiments 1 and 3 show that NRAR boosts the performance across most metrics, except for AP and $F_1$-max. (c) Comparing experiments 3 and 6, it can be seen that the combination of FM and NRAR further enhances performance across most metrics, except for AP and $F_1$-max. (d) Comparing experiments 1 and 4, it can be found that the segmentation network reduces AP and $F_1$-max. However, experiment 5 shows improvement when NRAR is added, indicating that the addition of pseudo-anomalous samples allows SN to improve performance across all metrics. Notably, AP and $F_1$-max gain significant improvement. SN can refine the predicted regions without affecting the distribution of predicted results. The best result is achieved by combining all three main designs.

\begin{table}[t]
  \centering
  \normalsize
  \caption{Ablations of CRR elements on Real-IAD (\%). FM: Feature Masking. NRAR: Normal Reconstruction and Anomaly Repair. SN: Segmentation Network.}
   \resizebox{\linewidth}{!}{
    \begin{tabular}{ccccccccccc}
    \toprule
    \multirow{2}[2]{*}{Exp.} & \multirow{2}[2]{*}{FM} & \multirow{2}[2]{*}{NRAR} & \multirow{2}[2]{*}{SN} & \multicolumn{3}{c}{Image-level} & \multicolumn{4}{c}{Pixel-level} \\
\cmidrule(r){5-7} \cmidrule(l){8-11} 
& & & & \multicolumn{1}{c}{AUROC} & \multicolumn{1}{c}{AP} & \multicolumn{1}{c}{$F_1$-max} & \multicolumn{1}{c}{AUROC} & \multicolumn{1}{c}{AP} & \multicolumn{1}{c}{$F_1$-max} & \multicolumn{1}{c}{AUPRO} \\ \midrule
1 &   &    &    & 89.33 & 86.77 & 80.17  & 98.84 & 42.79 & 47.10  & 93.86 \\
2 & \checkmark  &    &    & 89.56 & 87.10 & 80.31  & 98.88 & 44.35 & 48.34  & 94.66 \\
3 &   & \checkmark  &    & 90.98 & 89.27 & 82.16  & 98.93 & 39.26 & 45.21  & 94.87 \\
4 &   &   & \checkmark   & 89.34 & 86.77 & 80.22  & 98.98 & 39.13 & 44.56  & 94.25 \\
5 &   &  \checkmark  & \checkmark  & 91.09 & 89.63 & 82.16  & 99.06 & 51.26 & 52.11  & 95.57 \\
6 & \checkmark   &  \checkmark  &    & 90.99 & 89.33 & 81.96  & 99.07 & 37.08 & 43.40  & 94.98 \\
7 & \checkmark & \checkmark  &  \checkmark  & \textbf{91.32} & \textbf{89.67} & \textbf{82.63}  & \textbf{99.18} & \textbf{53.97} & \textbf{54.18}  & \textbf{95.79} \\
\bottomrule
\end{tabular}
}
\label{tab:abvisa}
\end{table}

\subsubsection{Mask rates}
\label{sec_ablation_mask_rates}
We performed ablation studies on the masking rate $\lambda$ in the MLP bottleneck after adding NRAR, as shown in Table \ref{tab4}. The experimental results demonstrate that CRR is robust to different levels of mask rates.
\begin{table}[t]
  \centering
  \tiny
  \caption{Ablations of Mask rates $\lambda$ in Bottleneck, conducted on Real-IAD (\%). \dag: default.}
   \resizebox{\linewidth}{!}{
    \begin{tabular}{lccccccc}
    \toprule
    \multirow{2}[2]{*}{$\lambda$} & \multicolumn{3}{c}{Image-level} & \multicolumn{4}{c}{Pixel-level} \\
\cmidrule(r){2-4} \cmidrule(l){5-8} 
 & \multicolumn{1}{c}{AUROC} & \multicolumn{1}{c}{AP} & \multicolumn{1}{c}{$F_1$-max} & \multicolumn{1}{c}{AUROC} & \multicolumn{1}{c}{AP} & \multicolumn{1}{c}{$F_1$-max} & \multicolumn{1}{c}{AUPRO} \\ \midrule
0    & 89.33 & 86.77 & 80.17  & 98.84 & \textbf{42.79} & \textbf{47.10}  & 93.86 \\
0.1    & 90.85 & 89.11 & 81.85  & 99.03 & 35.35 & 42.03  & 94.59 \\
0.2    & \textbf{91.01} & \textbf{89.37} & \textbf{82.06}  & 99.07 & 35.44 & 42.07 & 94.91 \\
0.3    & 90.84 & 88.95 & 81.75  & 99.06 & 36.21 & 42.91  & 94.89 \\
0.4 \dag & 90.99 & 89.33 & 81.96 & 99.07 & 37.08 & 43.40  & \textbf{94.98} \\
0.5    & 90.94 & 89.25 & 81.89 & \textbf{99.08} & 37.59 & 43.95 & 95.06 \\
0.6    & 90.15 & 88.35 & 80.88 & 99.03 & 37.95 & 44.05  & 94.67 \\
\bottomrule
\end{tabular}
}
\label{tab4}
\end{table}

\subsubsection{Segmentation framework}
As mentioned in Sec. \ref{SN-}, the segmentation network consists of four convolutional blocks and four upsampling modules. To validate the rationale of this setting, we compare it against a structure consisting of ResNet and Head~\cite{zhang2023destseg}. Both models receive the same input dimensions of $X$ without upsampling. We present the results in Tab. \ref{tab_seg}. It can be observed that our segmentation framework outperforms the ResNet-Head model, significantly improving the pixel-level AP and $F_1$-max.
\begin{table}[t]
  \centering
  \scriptsize
  \caption{Ablations of Segmentation framework, conducted on Real-IAD (\%).}
   \resizebox{\linewidth}{!}{
    \begin{tabular}{cccccccc}
    \toprule
    \multirow{2}[2]{*}{SN} & \multicolumn{3}{c}{Image-level} & \multicolumn{4}{c}{Pixel-level} \\
\cmidrule(r){2-4} \cmidrule(l){5-8} 
 & \multicolumn{1}{c}{AUROC} & \multicolumn{1}{c}{AP} & \multicolumn{1}{c}{$F_1$-max} & \multicolumn{1}{c}{AUROC} & \multicolumn{1}{c}{AP} & \multicolumn{1}{c}{$F_1$-max} & \multicolumn{1}{c}{AUPRO} \\ \midrule
ResNet-Head    & 86.1 & 85.0 & 77.6  & 96.1 & 46.2 & 54.2  & 89.0 \\
Conv-Upsample    & \textbf{89.2} & \textbf{87.9} & \textbf{80.2}  & \textbf{97.8} & \textbf{53.4} & \textbf{54.0} & \textbf{91.1} \\
\bottomrule
\end{tabular}
}
\label{tab_seg}
\end{table}
\subsection{Real-world applications}
\begin{figure}[t]
	\centering
	\includegraphics[width=\linewidth]{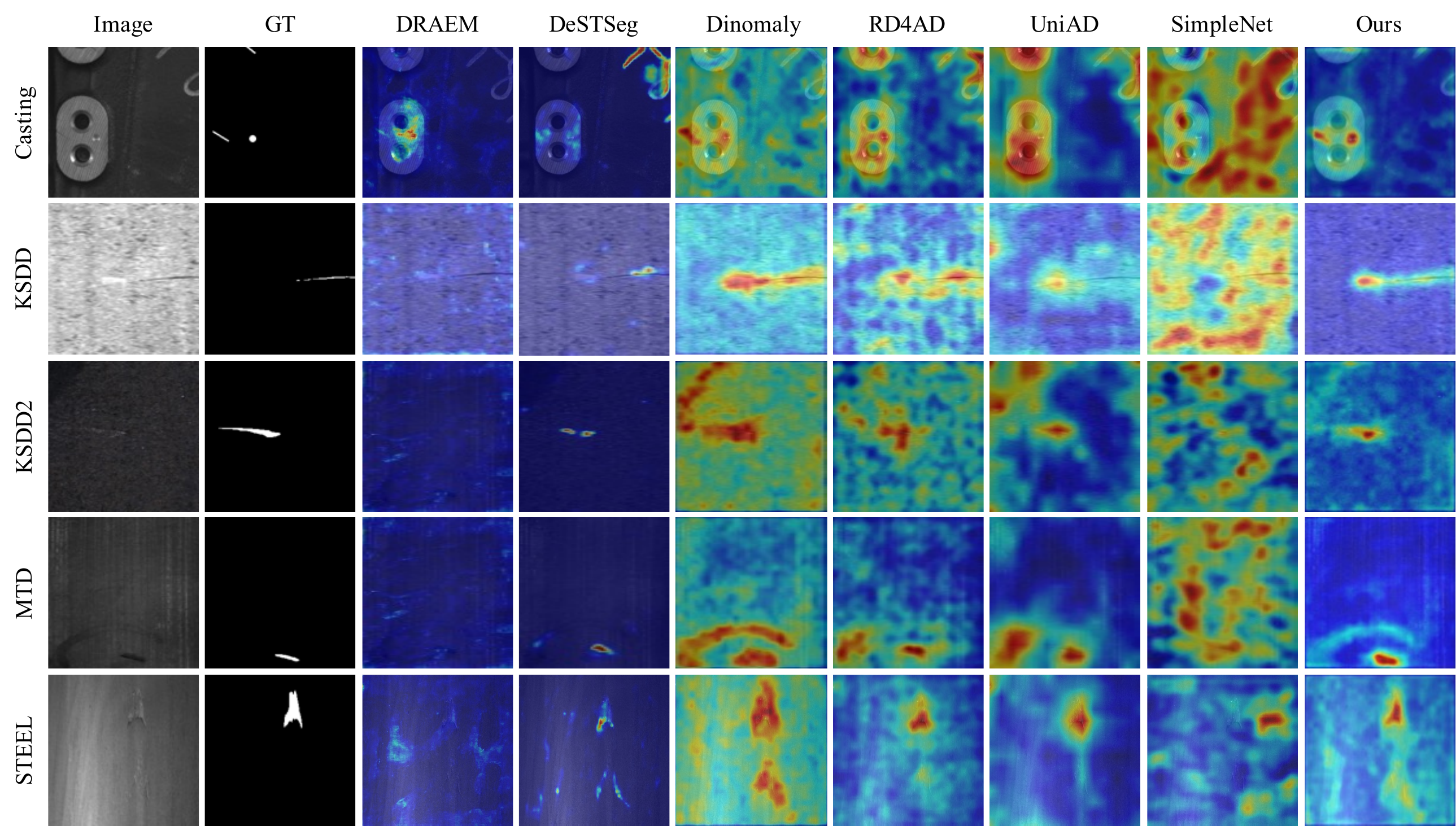}
	\caption{Qualitative anomaly localization results of the proposed model and comparative methods on the HSS-IAD dataset.}
	\label{fig:visual_hisiad}
\end{figure}
To further evaluate the applicability and generalization of the proposed CRR, we apply it to the real-world HSS-IAD~\cite{wang2025hss} dataset for multi-category surface defect detection in industrial products.
%

In Table~\ref{tab:results}, quantitative comparisons results on HSS-IAD are presented. CRR achieved image-level and pixel-level performances of \textbf{80.4}/\textbf{83.3}/\textbf{81.2} and \textbf{88.8}/\textbf{24.2}/\textbf{29.1}/\textbf{64.9}, respectively, showing improvements of \textbf{2.7}/\textbf{3.4}/\textbf{0.1} and \textbf{4.8}/\textbf{1.7}/\textbf{3.4}/\textbf{6.8}.
The qualitative anomaly localization comparison results of various methods are shown in Fig. \ref{fig:visual_hisiad}.
It is evident that our CRR method demonstrates an impressive capability to accurately locate anomalies that are easily confused with surface machining marks and process features, that are very similar to the material, that are tiny, and that have highly complex defect shapes.
%
%

\subsection{Discussion and limitations}

While our experiments sufficiently validate the efficacy of the proposed CRR, we acknowledge certain limitations. 
Notably, challenges emerge when dealing with logical defects, as evidenced by the $\sim$50\% AP on the Transistor dataset. 
Logical anomalies typically arise from violations of global or inter-component logical constraints, such as part existence, count, topology, or relative position, rather than pixel-level appearance changes.
Consequently, locally plausible yet globally inconsistent patterns may yield high image-level detection accuracy but poor localization, as pixel-wise segmentation struggles to delineate which pixels are anomalous.
This is consistent with prior reports~\cite{bergmann2022beyond} that reconstruction-centric pipelines underperform when anomalies are relational rather than appearance-based. 
In future investigations, we will develop component-aware and relation-consistency modeling to better detect logical anomalies.

%% file: sec/5_conclusion.tex
\section{CONCLUSION}
This paper proposes the CRR method, a collaborative reconstruction and repair framework to address the identity mapping issue in the MIAD task.
The decoder, endowed with the ability to reconstruct normal features while repairing abnormal features, is adopted to generate distinct features for anomalous regions and consistent features for normal areas.
Feature-level random masking is employed to restore or repair normal fine-grained feature representations, while a segmentation network is utilized to filter out inherent discrepancies between feature representations from encoder and decoder.
Extensive experiments on MVTec AD, VisA, Real-IAD, and HSS-IAD demonstrate the superiority of our approach over previous class-separated and unified multi-class models, highlighting the feasibility of implementing a unified model in complex real-world industrial scenarios. In future research endeavors, we intend to deploy this framework to a broader spectrum of industrial applications.

%% file: sec/6_appendix.tex
\appendix


\setcounter{table}{0}
\renewcommand{\thetable}{A\arabic{table}}

\setcounter{figure}{0}
\renewcommand{\thefigure}{A\arabic{figure}}

\section{Results Per-Category}
\label{sec:perclass}

\par For a more detailed analysis, we report the per-class results of MVTec-AD~\cite{bergmann2019mvtec}, VisA~\cite{zou2022visa}, Real-IAD~\cite{wang2024real}, and HSS-IAD~\cite{wang2025hss} from proposed CRR and compared methods. 
The results of image-level anomaly detection and pixel-level anomaly localization on MVTec-AD are presented in Table~\ref{tab:mvtecsp} and Table~\ref{tab:mvtecpx}, respectively. The results of image-level anomaly detection and pixel-level anomaly localization on VisA are presented in Table~\ref{tab:visasp} and Table~\ref{tab:visapx}, respectively. The results of image-level anomaly detection and pixel-level anomaly localization on Real-IAD are presented in Table~\ref{tab:realiadsp} and Table~\ref{tab:realiadpx}, respectively. The results of image-level anomaly detection and pixel-level anomaly localization on HSS-IAD are presented in Table~\ref{tab:hssiadsp} and Table~\ref{tab:hssiad_px}, respectively.
\par From these results, it can be concluded that CRR achieves state-of-the-art (SOTA) performance in almost all metrics across the majority of subcategories. However, CRR's anomaly classification performance in some subcategories of MVTec-AD and VisA did not show improvement compared to Dinomaly, primarily due to the near-saturation classification performance in these subcategories, which makes it difficult to highlight differences and advantages between methods. In contrast, there remains significant room for improvement in anomaly detection performance in the Real-IAD subcategories, where there is a greater disparity in performance between methods. It is evident that the proposed CRR method yields strong detection results across all subcategories of the Real-IAD. CRR performs well in image-level anomaly detection on the HSS-IAD dataset, but its performance on pixel-level localization metrics (AP / F1-max / AUPRO) is suboptimal, particularly for castings. This highlights the considerable practical challenges the model still faces in achieving precise localization.

\section{Qualitative Visualization}
\label{sec:visual}
\begin{figure}[!ht]
\centerline{\includegraphics[width=\linewidth]{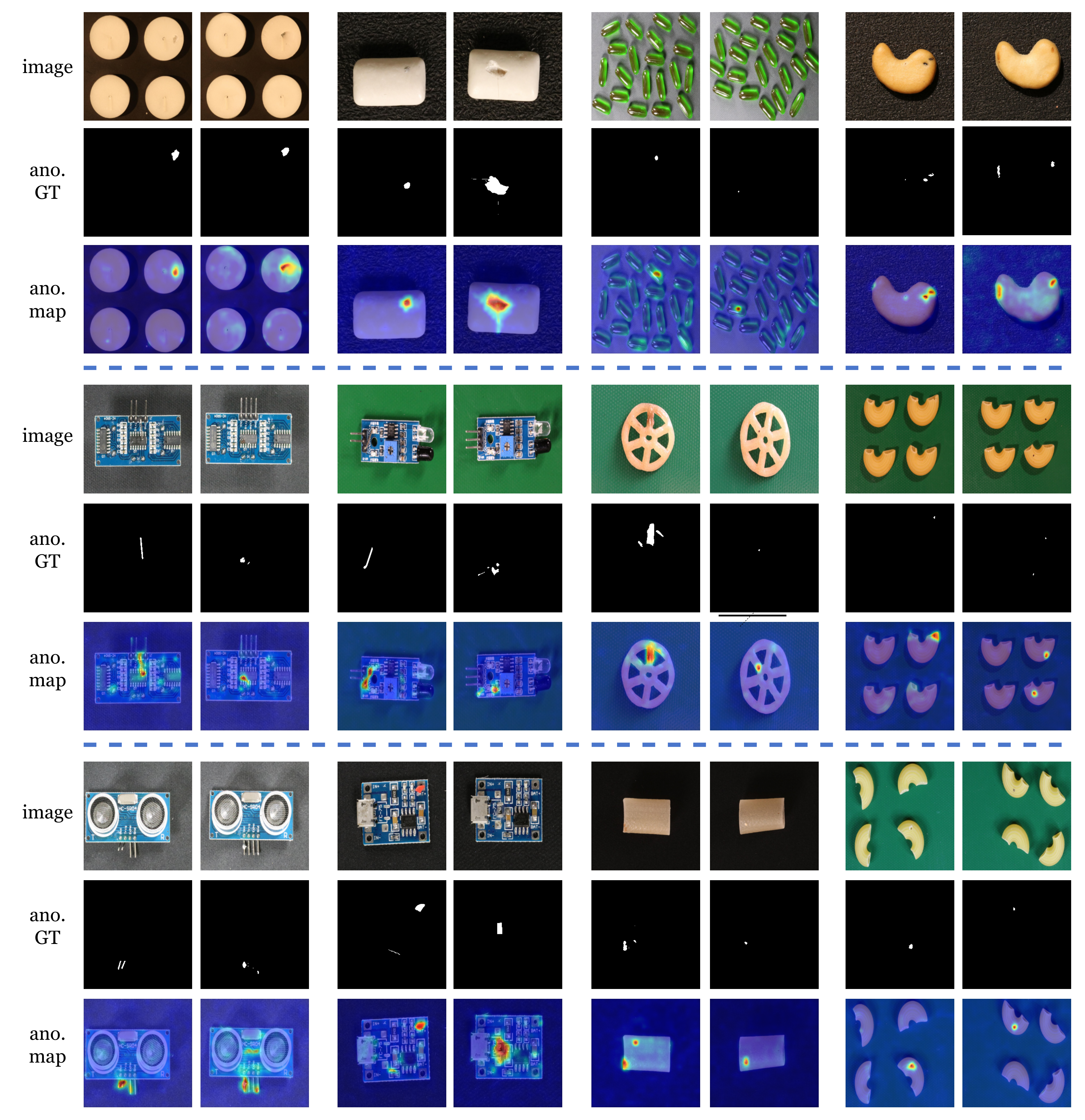}}
\caption{Anomaly maps visualization on VisA. All samples are randomly chosen.}
\label{fig_visa}
\end{figure}

\begin{figure}[!ht]
\centerline{\includegraphics[width=\linewidth]{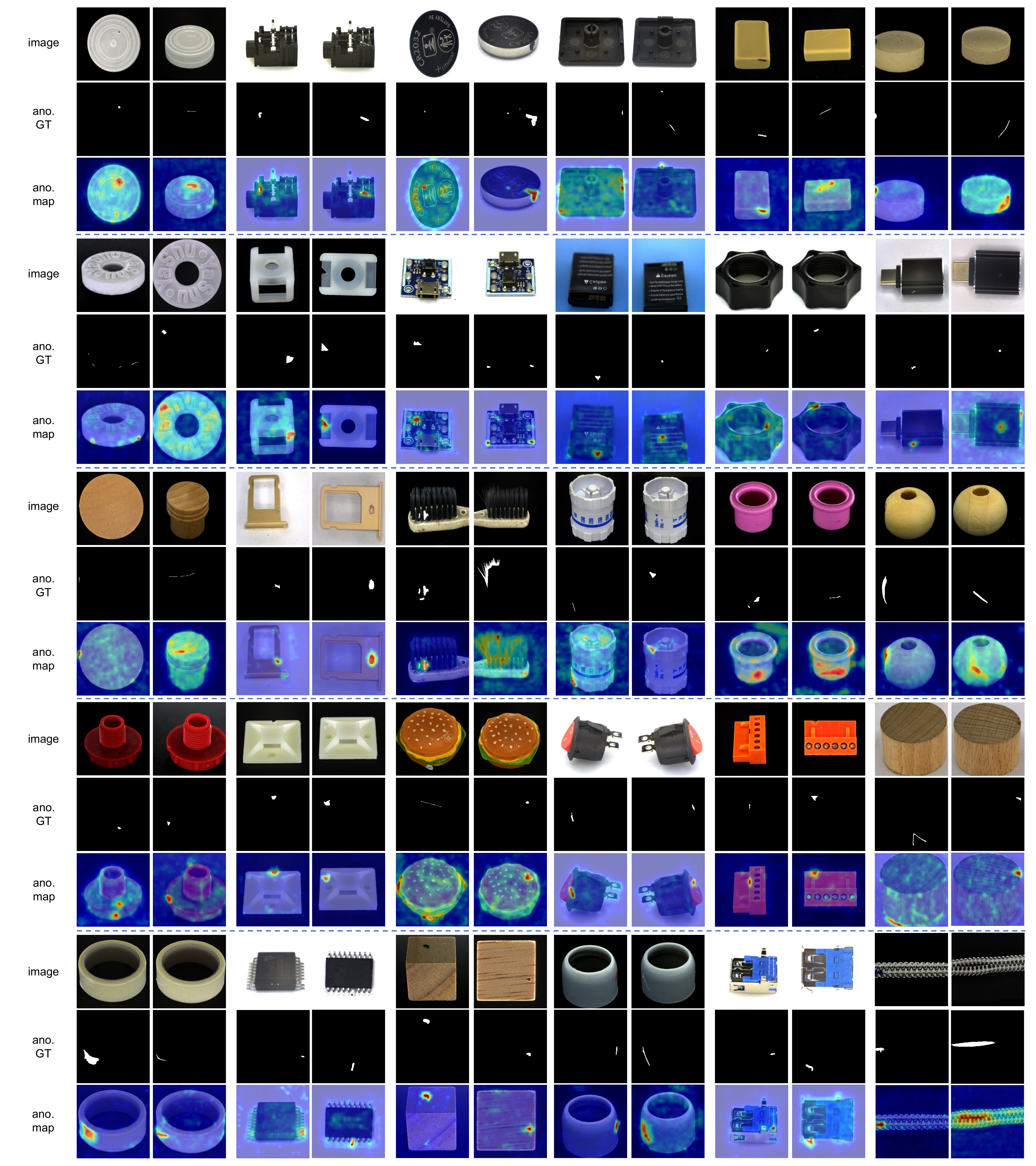}}
\caption{Anomaly maps visualization on Real-IAD. All samples are randomly chosen.}
\label{fig_realiad}
\end{figure}

\begin{figure}[!t]
\centerline{\includegraphics[width=0.82\linewidth]{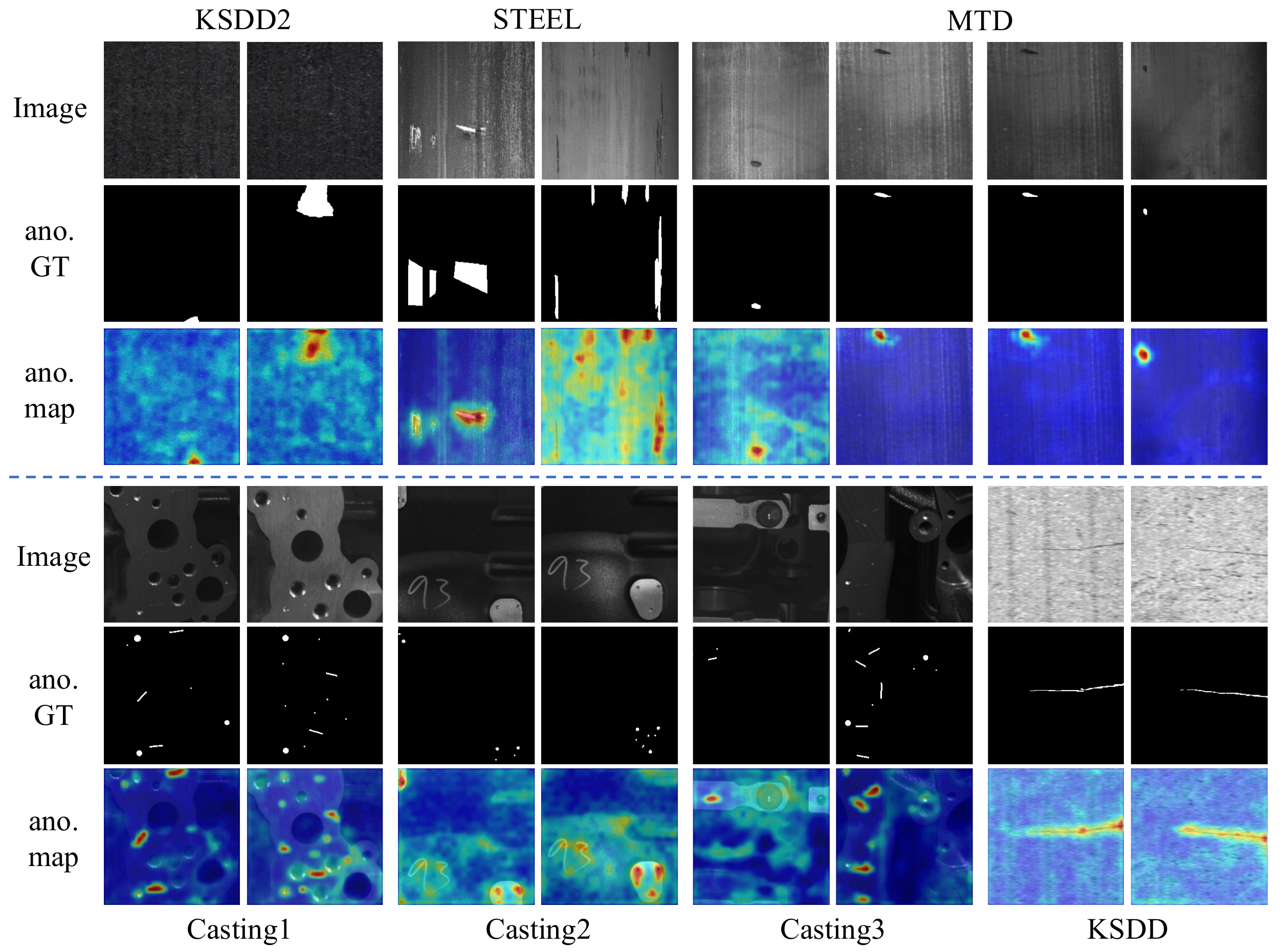}}
\caption{Anomaly maps visualization on HSS-IAD. All samples are randomly chosen.}
\label{fig_hssiad}
\end{figure}

\begin{figure}[!ht]
\centerline{\includegraphics[width=0.82\linewidth]{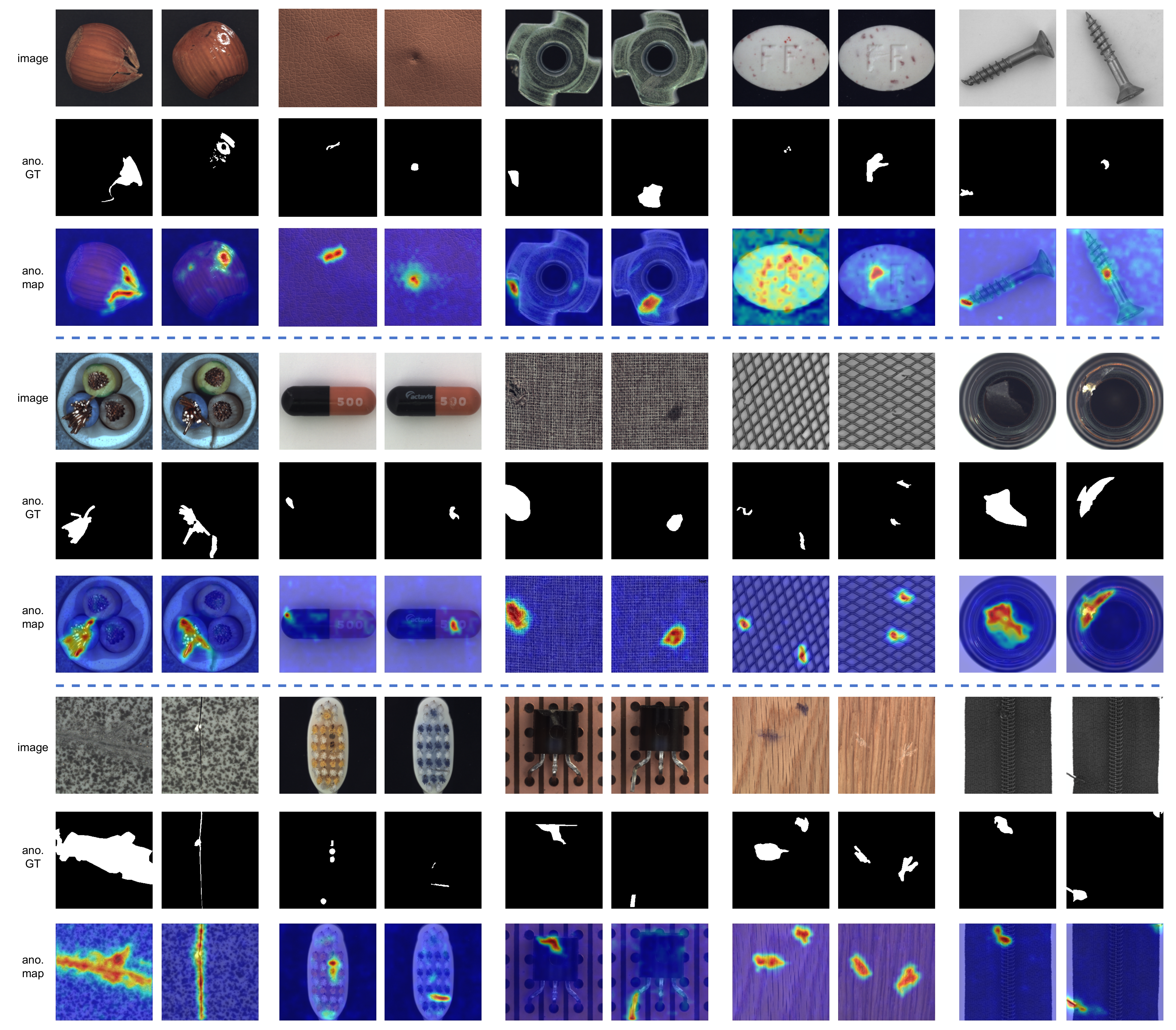}}
\caption{Anomaly maps visualization on MVTec-AD. All samples are randomly chosen.}
\label{fig_mvtec}
\end{figure}

We visualize the output anomaly maps of CRR on MVTec-AD, VisA, Real-IAD, and HSS-IAD, as shown in Figure~\ref{fig_mvtec}, Figure~\ref{fig_visa}, Figure~\ref{fig_realiad}, and Figure~\ref{fig_hssiad}. Note that all visualized samples were randomly selected without any artificial bias.

It can be observed that CRR accurately identifies and locates surface anomalies, whether they are minor scratches, small dents, or other subtle defects. Furthermore, it provides stable and precise anomaly localization results on structurally complex parts, further emphasizing the method's strong cross-category generalization capability. This makes it well-suited for using a unified model for anomaly detection across various industrial products.

\begin{table}[!ht]
  \centering
  \caption{Per-class performance on\textbf{ MVTec-AD} dataset for multi-class anomaly detection with AUROC/AP/$F_1$-max metrics.}
  \resizebox{1\linewidth}{!}{
    \begin{tabular}{p{3em}<{\centering} p{3.25em}<{\centering}p{6.2em}<{\centering} p{6.2em}<{\centering} p{6.2em}<{\centering} p{6.2em}<{\centering} p{6.2em}<{\centering} p{6.2em}<{\centering} p{6.2em}<{\centering} p{7em}<{\centering}}
    \toprule
    \multicolumn{2}{c}{Method~$\rightarrow$} & RD4AD~\cite{deng2022rd4ad} & UniAD~\cite{uniad} & SimpleNet~\cite{liu2023simplenet}  & DiAD~\cite{he2024diad} & Dinomaly~\cite{guo2024dinomaly} & \cellcolor{tab_ours}CRR \\
    \cline{1-2}
    \multicolumn{2}{c}{Category~$\downarrow$} & CVPR'22 & NeurlPS'22 & CVPR'23 & AAAI'24& Arxiv'24 & \cellcolor{tab_ours}Ours \\
    \hline
    \multicolumn{1}{c}{\multirow{10}[1]{*}{\begin{turn}{-90}Objects\end{turn}}} & \multicolumn{1}{c}{Bottle} & 99.6/99.9/98.4 & 99.7/\textbf{100.}/\textbf{100.} & \textbf{100.}/\textbf{100.}/\textbf{100.} & 99.7/96.5/91.8 &\textbf{100.}/\textbf{100.}/\textbf{100.} & \cellcolor{tab_ours}\textbf{100.}/\textbf{100.}/\textbf{100.} \\
    
    \multicolumn{1}{c}{} & \multicolumn{1}{c}{\cellcolor{tab_others}Cable} & \cellcolor{tab_others}84.1/89.5/82.5 & \cellcolor{tab_others}95.2/95.9/88.0 & \cellcolor{tab_others}97.5/98.5/94.7 & \cellcolor{tab_others}94.8/98.8/95.2 &\cellcolor{tab_others}\textbf{100.}/\textbf{100.}/\textbf{100.} & \cellcolor{tab_ours}99.6/99.8/98.3 \\
    
    \multicolumn{1}{c}{} & \multicolumn{1}{c}{Capsule}  & 94.1/96.9/96.9 & 86.9/97.8/94.4 & 90.7/97.9/93.5 & 89.0/97.5/95.5 &97.9/99.5/97.7 & \cellcolor{tab_ours}\textbf{98.4}/\textbf{99.6}/\textbf{98.6}\\
    
    \multicolumn{1}{c}{} & \multicolumn{1}{c}{\cellcolor{tab_others}Hazelnut} & \cellcolor{tab_others}60.8/69.8/86.4 & \cellcolor{tab_others}99.8/\textbf{100.}/99.3 & \cellcolor{tab_others}99.9/99.9/99.3 & \cellcolor{tab_others}99.5/99.7/97.3 &\cellcolor{tab_others}\textbf{100.}/\textbf{100.}/\textbf{100.} & \cellcolor{tab_ours}\textbf{100.}/\textbf{100.}/\textbf{100.}\\
    
    \multicolumn{1}{c}{} & \multicolumn{1}{c}{Metal Nut}  & \textbf{100.}/\textbf{100.}/99.5 & 99.2/99.9/99.5 & 96.9/99.3/96.1 & 99.1/96.0/91.6 &\textbf{100.}/\textbf{100.}/\textbf{100.} & \cellcolor{tab_ours}\textbf{100.}/\textbf{100.}/\textbf{100.} \\
    
    \multicolumn{1}{c}{} & \multicolumn{1}{c}{\cellcolor{tab_others}Pill}  & \cellcolor{tab_others}97.5/99.6/96.8 & \cellcolor{tab_others}93.7/98.7/95.7 & \cellcolor{tab_others}88.2/97.7/92.5 & \cellcolor{tab_others}95.7/98.5/94.5 &\cellcolor{tab_others}99.1/\textbf{99.9}/98.3 & \cellcolor{tab_ours}\textbf{99.3}/\textbf{99.9}/\textbf{98.6}\\
    
    \multicolumn{1}{c}{} & \multicolumn{1}{c}{Screw} & 97.7/99.3/95.8 & 87.5/96.5/89.0 & 76.7/90.6/87.7 & 90.7/\textbf{99.7}/\textbf{97.9} & 98.4/99.5/96.1 & \cellcolor{tab_ours}\textbf{99.0}/\textbf{99.7}/97.5\\
    
    \multicolumn{1}{c}{} & \multicolumn{1}{c}{\cellcolor{tab_others}Toothbrush} & \cellcolor{tab_others}97.2/99.0/94.7 & \cellcolor{tab_others}94.2/97.4/95.2& \cellcolor{tab_others}89.7/95.7/92.3 & \cellcolor{tab_others}99.7/99.9/99.2&\cellcolor{tab_others}\textbf{100.}/\textbf{100.}/\textbf{100.} & \cellcolor{tab_ours}\textbf{100.}/\textbf{100.}/\textbf{100.} \\
    
    \multicolumn{1}{c}{} & \multicolumn{1}{c}{Transistor} & 94.2/95.2/90.0 & 99.8/98.0/93.8 & 99.2/98.7/\textbf{97.6} & \textbf{99.8}/\textbf{99.6}/97.4 &99.0/98.0/96.4 & \cellcolor{tab_ours}99.5/99.2/\textbf{97.6}\\
    
    \multicolumn{1}{c}{} & \multicolumn{1}{c}{\cellcolor{tab_others}Zipper}  & \cellcolor{tab_others}99.5/99.9/99.2 & \cellcolor{tab_others}95.8/99.5/97.1 & \cellcolor{tab_others}99.0/99.7/98.3 & \cellcolor{tab_others}95.1/99.1/94.4 &\cellcolor{tab_others}\textbf{100.}/\textbf{100.}/\textbf{100.} & \cellcolor{tab_ours}\textbf{100.}/\textbf{100.}/\textbf{100.}\\
    \hline
    \multicolumn{1}{c}{\multirow{5}[1]{*}{\begin{turn}{-90}Textures\end{turn}}} & \multicolumn{1}{c}{Carpet}  & 98.5/99.6/97.2 & \textbf{99.8}/99.9/99.4 & 95.7/98.7/93.2 & 99.4/99.9/98.3 &\textbf{99.8}/\textbf{100.}/98.9 & \cellcolor{tab_ours}\textbf{99.8}/\textbf{100.}/\textbf{99.5}\\
    
    \multicolumn{1}{c}{} & \multicolumn{1}{c}{\cellcolor{tab_others}Grid} & \cellcolor{tab_others}98.0/99.4/96.5 & \cellcolor{tab_others}98.2/99.5/97.3 & \cellcolor{tab_others}97.6/99.2/96.4 & \cellcolor{tab_others}98.5/99.8/97.7 & \textbf{99.9}/\textbf{100.}/\textbf{99.1} & \cellcolor{tab_ours}99.8/99.9/\textbf{99.1}\\
    
    \multicolumn{1}{c}{} & \multicolumn{1}{c}{Leather} & \textbf{100.}/\textbf{100.}/\textbf{100.} & \textbf{100.}/\textbf{100.}/\textbf{100.} & \textbf{100.}/\textbf{100.}/\textbf{100.} & 99.8/99.7/97.6 &\textbf{100.}/\textbf{100.}/\textbf{100.} & \cellcolor{tab_ours}\textbf{100.}/\textbf{100.}/\textbf{100.}\\
    
    \multicolumn{1}{c}{} & \multicolumn{1}{c}{\cellcolor{tab_others}Tile}  & \cellcolor{tab_others}98.3/99.3/96.4 & \cellcolor{tab_others}99.3/99.8/98.2 & \cellcolor{tab_others}99.3/99.8/98.8 & \cellcolor{tab_others}96.8/99.9/98.4 &\cellcolor{tab_others}\textbf{100.}/\textbf{100.}/\textbf{100.} & \cellcolor{tab_ours}\textbf{100.}/\textbf{100.}/\textbf{100.}\\
    
    \multicolumn{1}{c}{} & \multicolumn{1}{c}{Wood}  & 99.2/99.8/98.3 & 98.6/99.6/96.6 & 98.4/99.5/96.7 & 99.7/\textbf{100.}/\textbf{100.} &\textbf{99.8}/99.9/99.2 & \cellcolor{tab_ours}\textbf{99.8}/99.9/99.2\\
    \hline
    \multicolumn{2}{c}{\cellcolor{tab_others}Mean} & \cellcolor{tab_others}94.6/96.5/95.2 & \cellcolor{tab_others}96.5/98.8/96.2 & \cellcolor{tab_others}95.3/98.4/95.8 & \cellcolor{tab_others}97.2/99.0/96.5 &\cellcolor{tab_others}99.6/99.8/99.0  &  \cellcolor{tab_ours}\textbf{99.7}/\textbf{99.9}/\textbf{99.2}\\
    \bottomrule
    \end{tabular}%
  }
  \label{tab:mvtecsp}%
\end{table}%

\begin{table}[!ht]
\centering
\caption{Per-class performance on \textbf{MVTec-AD} dataset for multi-class anomaly localization with AUROC/AP/$F_1$-max/AUPRO metrics.}
\resizebox{\linewidth}{!}{
\begin{tabular}{ccccccccc}
\toprule
\multicolumn{2}{c}{Method~$\rightarrow$} & RD4AD~\cite{deng2022rd4ad} & UniAD~\cite{uniad} & SimpleNet~\cite{liu2023simplenet} & DiAD~\cite{he2024diad} & Dinomaly~\cite{guo2024dinomaly} & \cellcolor{tab_ours}CRR \\
\cline{1-2}
\multicolumn{2}{c}{Category~$\downarrow$} & CVPR'22 & NeurlPS'22 & CVPR'23 & AAAI'24& Arxiv'24 & \cellcolor{tab_ours}Ours \\
\hline
\multicolumn{1}{c}{\multirow{10}[1]{*}{\begin{turn}{-90}Objects\end{turn}}} & \multicolumn{1}{c}{Bottle} & 97.8/68.2/67.6/94.0 & 98.1/66.0/69.2/93.1 & 97.2/53.8/62.4/89.0 & 98.4/52.2/54.8/86.6 & 99.2/88.6/84.2/96.6 & \cellcolor{tab_ours}\textbf{99.2}/\textbf{90.4}/\textbf{84.5}/\textbf{97.5} \\

\multicolumn{1}{c}{} & \multicolumn{1}{c}{\cellcolor{tab_others}Cable} & \cellcolor{tab_others}85.1/26.3/33.6/75.1 & \cellcolor{tab_others}97.3/39.9/45.2/86.1 & \cellcolor{tab_others}96.7/42.4/51.2/85.4 & \cellcolor{tab_others}96.8/50.1/57.8/80.5 & \cellcolor{tab_others}\textbf{98.6}/\textbf{72.0}/\textbf{74.3}/\textbf{94.2} & \cellcolor{tab_ours}98.5/71.0/70.1/94.1 \\

\multicolumn{1}{c}{} & \multicolumn{1}{c}{Capsule} & \textbf{98.8}/43.4/50.0/94.8 & 98.5/42.7/46.5/92.1 &98.5/35.4/44.3/84.5 & 97.1/42.0/45.3/87.2 & 98.7/61.4/\textbf{60.3}/97.2 & \textbf{98.9}/\cellcolor{tab_ours}\textbf{63.1}/59.5/\textbf{97.3} \\

\multicolumn{1}{c}{} & \multicolumn{1}{c}{\cellcolor{tab_others}Hazelnut} & \cellcolor{tab_others}97.9/36.2/51.6/92.7 & \cellcolor{tab_others}98.1/55.2/56.8/94.1 &\cellcolor{tab_others}98.4/44.6/51.4/87.4 & \cellcolor{tab_others}98.3/79.2/\textbf{80.4}/91.5& \cellcolor{tab_others}\textbf{99.4}/\textbf{82.2}/\textbf{76.4}/\textbf{97.0} & \cellcolor{tab_ours}99.3/77.3/74.8/96.7 \\

\multicolumn{1}{c}{} & \multicolumn{1}{c}{Metal Nut} & 94.8/55.5/66.4/91.9 & 62.7/14.6/29.2/81.8 &\textbf{98.0}/\textbf{83.1}/79.4/85.2 & 97.3/30.0/38.3/90.6 & 96.9/78.6/86.7/94.9 & \cellcolor{tab_ours}\textbf{97.4}/\textbf{83.5}/\textbf{87.0}/\textbf{96.5} \\

\multicolumn{1}{c}{} & \multicolumn{1}{c}{\cellcolor{tab_others}Pill} & \cellcolor{tab_others}97.5/63.4/65.2/95.8 & \cellcolor{tab_others}95.0/44.0/53.9/95.3 &\cellcolor{tab_others}96.5/72.4/67.7/81.9 & \cellcolor{tab_others}95.7/46.0/51.4/89.0 & \cellcolor{tab_others}\textbf{97.8}/\textbf{74.5}/\textbf{69.6}/\textbf{97.8} & \cellcolor{tab_ours}\textbf{97.8}/\textbf{74.5}/\textbf{69.6}/\textbf{97.8} \\

\multicolumn{1}{c}{} & \multicolumn{1}{c}{Screw} & 99.4/40.2/44.6/96.8 & 98.3/28.7/37.6/95.2 & 96.5/15.9/23.2/84.0 & 97.9/\textbf{60.6}/\textbf{59.6}/95.0 & 99.6/60.2/59.6/98.3 & \cellcolor{tab_ours}\textbf{99.7}/\textbf{63.0}/\textbf{60.1}/\textbf{98.7} \\

\multicolumn{1}{c}{} & \multicolumn{1}{c}{\cellcolor{tab_others}Toothbrush} & \cellcolor{tab_others}\textbf{99.0}/53.6/58.8/92.0 & \cellcolor{tab_others}98.4/34.9/45.7/87.9& \cellcolor{tab_others}98.4/46.9/52.5/87.4 & \cellcolor{tab_others}\textbf{99.0}/78.7/72.8/95.0 & \cellcolor{tab_others}98.9/51.5/62.6/95.3 & \cellcolor{tab_ours} \textbf{99.2}/\textbf{58.6}/\textbf{64.6}/\textbf{96.1} \\

\multicolumn{1}{c}{} & \multicolumn{1}{c}{Transistor} & 85.9/42.3/45.2/74.7 & \textbf{97.9}/59.5/\textbf{64.6}/\textbf{93.5} & 95.8/58.2/56.0/83.2 & 95.1/15.6/31.7/90.0 &\textbf{93.2}/\textbf{59.9}/\textbf{58.5}/77.0 & \cellcolor{tab_ours}93.0/58.7/56.9/\textbf{77.7} \\

\multicolumn{1}{c}{} & \multicolumn{1}{c}{\cellcolor{tab_others}Zipper} & \cellcolor{tab_others}98.5/53.9/60.3/94.1 &\cellcolor{tab_others} 96.8/40.1/49.9/92.6 & \cellcolor{tab_others}97.9/53.4/54.6/90.7 & \cellcolor{tab_others}96.2/60.7/60.0/91.6 & \cellcolor{tab_others}\textbf{99.2}/79.5/75.4/97.2 & \cellcolor{tab_ours}\textbf{99.2}/\textbf{81.3}/\textbf{76.3}/\textbf{97.7} \\
\hline

\multicolumn{1}{c}{\multirow{5}[1]{*}{\begin{turn}{-90}Textures\end{turn}}} & \multicolumn{1}{c}{Carpet} & 99.0/58.5/60.4/95.1 & 98.5/49.9/51.1/94.4 & 97.4/38.7/43.2/90.6 & 98.6/42.2/46.4/90.6 &99.3/68.7/71.1/97.6 & \cellcolor{tab_ours}\textbf{99.4}/\textbf{72.9}/\textbf{71.3}/\textbf{98.5} \\

\multicolumn{1}{c}{} & \multicolumn{1}{c}{\cellcolor{tab_others}Grid} & \cellcolor{tab_others}96.5/23.0/28.4/97.0 & \cellcolor{tab_others}63.1/10.7/11.9/92.9 & \cellcolor{tab_others}96.8/20.5/27.6/88.6/ & \cellcolor{tab_others}96.6/66.0/64.1/94.0 & \cellcolor{tab_others}99.4/55.3/57.7/97.2 & \cellcolor{tab_ours}\textbf{99.6}/\textbf{64.8}/\textbf{62.4}/\textbf{98.3}\\

\multicolumn{1}{c}{} & \multicolumn{1}{c}{Leather} & 99.3/38.0/45.1/97.4 & 98.8/32.9/34.4/96.8 &98.7/28.5/32.9/92.7 & 98.8/56.1/62.3/91.3 & 99.4/52.2/55.0/97.6 & \cellcolor{tab_ours}\textbf{99.4}/\textbf{61.0}/\textbf{55.5}/\textbf{98.9} \\

\multicolumn{1}{c}{} & \multicolumn{1}{c}{\cellcolor{tab_others}Tile} & \cellcolor{tab_others}95.3/48.5/60.5/85.8 & \cellcolor{tab_others}91.8/42.1/50.6/78.4 & \cellcolor{tab_others}95.7/60.5/59.9/90.6 & \cellcolor{tab_others}92.4/65.7/64.1/\textbf{90.7} & \cellcolor{tab_others}98.1/80.1/75.7/90.5 & \cellcolor{tab_ours}97.9/79.0/73.6/\textbf{90.7} \\

\multicolumn{1}{c}{} & \multicolumn{1}{c}{Wood} & 95.3/47.8/51.0/90.0 & 93.2/37.2/41.5/86.7 &91.4/34.8/39.7/76.3 & 93.3/43.3/43.5/\textbf{97.5} & \textbf{97.6}/\textbf{72.8}/\textbf{68.4}/94.0 & \cellcolor{tab_ours}97.4/72.0/67.3/\textbf{95.3} \\
\hline

\multicolumn{2}{c}{\cellcolor{tab_others}Mean} & \cellcolor{tab_others}96.1/48.6/53.8/91.1 & \cellcolor{tab_others}96.8/43.4/49.5/90.7 & \cellcolor{tab_others}96.9/45.9/49.7/86.5 & \cellcolor{tab_others}96.8/52.6/55.5/90.7 & \cellcolor{tab_others}\textbf{98.4}/69.3/\textbf{69.2}/94.8 & \cellcolor{tab_ours}\textbf{98.4}/\textbf{71.4}/68.9/\textbf{95.5} \\

\hline
\multicolumn{2}{c}{mAD} &82.3  &81.7  &81.2  &84.0  & 90.0 & \cellcolor{tab_ours}{\textbf{90.4}} \\
\bottomrule
\end{tabular}
}
\label{tab:mvtecpx}%
\end{table}%
\begin{table}[!ht]
\centering
\caption{Per-class performance on \textbf{VisA} dataset for multi-class anomaly detection with AUROC/AP/$F_1$-max metrics.}
\resizebox{1.\linewidth}{!}{
\begin{tabular}{cccccccc}
\toprule
Method~$\rightarrow$ & RD4AD~\cite{deng2022rd4ad} & UniAD~\cite{uniad} & SimpleNet~\cite{liu2023simplenet} & DiAD~\cite{he2024diad} & Dinomaly~\cite{guo2024dinomaly} & \cellcolor{tab_ours}CRR \\
\cline{1-1}
Category~$\downarrow$ & CVPR'22 & NeurlPS'22 & CVPR'23 & AAAI'24& Arxiv'24 & \cellcolor{tab_ours}Ours \\
\hline

pcb1 & 96.2/95.5/91.9 & 92.8/92.7/87.8 & 91.6/91.9/86.0 & 88.1/88.7/80.7 & \textbf{99.1}/\textbf{99.1}/\textbf{96.6} & \cellcolor{tab_ours}\textbf{99.1}/\textbf{99.1}/96.0 \\

\cellcolor{tab_others}pcb2 & \cellcolor{tab_others}97.8/97.8/94.2 & \cellcolor{tab_others}87.8/87.7/83.1 &\cellcolor{tab_others}92.4/93.3/84.5 & \cellcolor{tab_others}91.4/91.4/84.7 &\cellcolor{tab_others}99.3/99.2/97.0 & \cellcolor{tab_ours}\textbf{99.7}/\textbf{99.7}/\textbf{98.5} \\

pcb3 & 96.4/96.2/91.0 & 78.6/78.6/76.1 & 89.1/91.1/82.6 & 86.2/87.6/77.6 & \textbf{98.9}/\textbf{98.9}/\textbf{96.1} & \cellcolor{tab_ours}\textbf{98.9}/\textbf{98.9}/94.5 \\

\cellcolor{tab_others}pcb4 & \cellcolor{tab_others}99.9/99.9/99.0 & \cellcolor{tab_others}98.8/98.8/94.3 & \cellcolor{tab_others}97.0/97.0/93.5 &\cellcolor{tab_others}99.6/99.5/97.0 &\cellcolor{tab_others}\textbf{99.8}/\textbf{99.8}/98.0 & \cellcolor{tab_ours}\textbf{99.8}/\textbf{99.8}/\textbf{98.5} \\
\hline

macaroni1 & 75.9/ 1.5/76.8 & 79.9/79.8/72.7 & 85.9/82.5/73.1 & 85.7/85.2/78.8 & 98.0/97.6/94.2 & \cellcolor{tab_ours}\textbf{99.5}/\textbf{99.3}/\textbf{99.0} \\

\cellcolor{tab_others}macaroni2 & \cellcolor{tab_others}88.3/84.5/83.8 & \cellcolor{tab_others}71.6/71.6/69.9 &\cellcolor{tab_others}68.3/54.3/59.7 & \cellcolor{tab_others}62.5/57.4/69.6 &\cellcolor{tab_others}95.9/95.7/90.7 & \cellcolor{tab_ours}\textbf{98.4}/\textbf{98.4}/\textbf{94.1} \\

capsules & 82.2/90.4/81.3 & 55.6/55.6/76.9 &74.1/82.8/74.6 & 58.2/69.0/78.5 & 98.6/99.0/97.1 & \cellcolor{tab_ours}\textbf{99.5}/\textbf{99.7}/\textbf{99.0} \\

\cellcolor{tab_others}candle & \cellcolor{tab_others}92.3/92.9/86.0 & \cellcolor{tab_others}94.1/94.0/86.1 & \cellcolor{tab_others}84.1/73.3/76.6 & \cellcolor{tab_others}92.8/92.0/87.6 &\cellcolor{tab_others}\textbf{98.7}/\textbf{98.8}/\textbf{95.1} & \cellcolor{tab_ours}97.6/97.7/92.5 \\
\hline

cashew & 92.0/95.8/90.7 & 92.8/92.8/91.4 & 88.0/91.3/84.7 & 91.5/95.7/89.7 & 98.7/\textbf{99.4}/\textbf{97.0} & \cellcolor{tab_ours}\textbf{98.8}/\textbf{99.4}/96.5 \\

\cellcolor{tab_others}chewinggum &\cellcolor{tab_others}94.9/97.5/92.1 &\cellcolor{tab_others}96.3/96.2/95.2 & \cellcolor{tab_others}96.4/98.2/93.8 & \cellcolor{tab_others}99.1/99.5/95.9 &\cellcolor{tab_others}\textbf{99.8}/\textbf{99.9}/\textbf{99.0} & \cellcolor{tab_ours}\textbf{99.8}/\textbf{99.9}/\textbf{98.5} \\

fryum & 95.3/97.9/91.5 & 83.0/83.0/85.0 & 88.4/93.0/83.3 & 89.8/95.0/87.2 & 98.8/99.4/96.5 & \cellcolor{tab_ours}\textbf{99.4}/\textbf{99.7}/\textbf{97.4} \\

\cellcolor{tab_others}pipe\_fryum &\cellcolor{tab_others}97.9/98.9/96.5 & \cellcolor{tab_others}94.7/94.7/93.9 &\cellcolor{tab_others}90.8/95.5/88.6 & \cellcolor{tab_others}96.2/98.1/93.7 &\cellcolor{tab_others}99.2/99.7/97.0 & \cellcolor{tab_ours}\textbf{99.9}/\textbf{100.}/\textbf{99.5} \\
\hline

Mean  & 92.4/92.4/89.6   & 85.5/85.5/84.4 & 87.2/87.0/81.8 & 86.8/88.3/85.1& 98.7/98.9/96.2 &   \cellcolor{tab_ours}\textbf{99.2}/\textbf{99.3}/\textbf{97.0} \\
\bottomrule
\end{tabular}
  }
  \label{tab:visasp}%
\end{table}%
\begin{table}[!ht]
\centering
\caption{Per-class performance on \textbf{VisA} dataset for multi-class anomaly localization with AUROC/AP/$F_1$-max/AUPRO metrics.}
\resizebox{1\linewidth}{!}{
\begin{tabular}{cccccccc}
\toprule
Method~$\rightarrow$ & RD4AD~\cite{deng2022rd4ad} & UniAD~\cite{uniad} & SimpleNet~\cite{liu2023simplenet} & DiAD~\cite{he2024diad} & Dinomaly~\cite{guo2024dinomaly} & \cellcolor{tab_ours}CRR \\
\cline{1-1}
Category~$\downarrow$ & CVPR'22 & NeurlPS'22 & CVPR'23 & AAAI'24& Arxiv'24& \cellcolor{tab_ours} Ours \\
\midrule

pcb1 & 99.4/66.2/62.4/\textbf{95.8} & 93.3/ 3.9/ 8.3/64.1 & 99.2/86.1/78.8/83.6 & 98.7/49.6/52.8/80.2 &\textbf{99.5}/87.9/80.5/95.1 &\cellcolor{tab_ours}99.3/\textbf{89.5}/\textbf{82.5}/\textbf{95.4} \\

\cellcolor{tab_others}pcb2 & \cellcolor{tab_others}98.0/22.3/30.0/90.8 & \cellcolor{tab_others}93.9/ 4.2/ 9.2/66.9 & \cellcolor{tab_others}96.6/ 8.9/18.6/85.7 & \cellcolor{tab_others}95.2/ 7.5/16.7/67.0 &\cellcolor{tab_others}\textbf{98.0}/47.0/49.8/91.3 &\cellcolor{tab_ours}97.0/\textbf{48.3}/\textbf{50.9}/\textbf{91.7}\\

pcb3 & 97.9/26.2/35.2/93.9 & 97.3/13.8/21.9/70.6 & 97.2/31.0/36.1/85.1 & 96.7/ 8.0/18.8/68.9 &\textbf{98.4}/\textbf{41.7}/\textbf{45.3}/\textbf{94.6} &\cellcolor{tab_ours}96.3/38.6/41.9/94.2\\

\cellcolor{tab_others}pcb4 &\cellcolor{tab_others}97.8/31.4/37.0/88.7 & \cellcolor{tab_others}94.9/14.7/22.9/72.3 &\cellcolor{tab_others}93.9/23.9/32.9/61.1 & \cellcolor{tab_others}97.0/17.6/27.2/85.0 &\cellcolor{tab_others}\textbf{98.7}/\textbf{50.5}/\textbf{53.1}/\textbf{94.4} &\cellcolor{tab_ours}97.7/47.1/50.5/90.9\\

\midrule
macaroni1 & 99.4/ 2.9/6.9/95.3 & 97.4/ 3.7/ 9.7/84.0 & 98.9/ 3.5/8.4/92.0 & 94.1/10.2/16.7/68.5 &99.6/33.5/\textbf{40.6}/96.4 &\cellcolor{tab_ours}\textbf{99.9}/\textbf{34.9}/37.1/\textbf{99.5}\\

\cellcolor{tab_others}macaroni2 & \cellcolor{tab_others}99.7/13.2/21.8/97.4 & \cellcolor{tab_others}95.2/ 0.9/ 4.3/76.6 & \cellcolor{tab_others}93.2/ 0.6/ 3.9/77.8 & \cellcolor{tab_others}93.6/ 0.9/ 2.8/73.1 &\cellcolor{tab_others}99.7/24.7/\textbf{36.1}/98.7 &\cellcolor{tab_ours}\textbf{99.9}/\textbf{25.0}/32.2/\textbf{99.5}\\

capsules &99.4/60.4/60.8/93.1 & 88.7/ 3.0/ 7.4/43.7 &97.1/52.9/53.3/73.7 & 97.3/10.0/21.0/77.9 &99.6/65.0/66.6/97.4 &\cellcolor{tab_ours}\textbf{99.8}/\textbf{73.5}/\textbf{71.3}/\textbf{99.2}\\

\cellcolor{tab_others}candle & \cellcolor{tab_others}99.1/25.3/35.8/94.9 & \cellcolor{tab_others}98.5/17.6/27.9/91.6 & \cellcolor{tab_others}97.6/ 8.4/16.5/87.6 & \cellcolor{tab_others}97.3/12.8/22.8/89.4 &\cellcolor{tab_others}99.4/43.0/47.9/95.4 &\cellcolor{tab_ours}\textbf{99.7}/\textbf{43.8}/\textbf{51.1}/\textbf{98.8}\\

\midrule
cashew & 91.7/44.2/49.7/86.2 & 98.6/51.7/58.3/87.9 & 98.9/\textbf{68.9}/66.0/84.1 & 90.9/53.1/60.9/61.8 &97.1/64.5/62.4/94.0 &\cellcolor{tab_ours}\textbf{99.3}/68.1/\textbf{66.2}/\textbf{98.4}\\

\cellcolor{tab_others}chewinggum & \cellcolor{tab_others}98.7/59.9/61.7/76.9 & \cellcolor{tab_others}98.8/54.9/56.1/81.3 & \cellcolor{tab_others}97.9/26.8/29.8/78.3 & \cellcolor{tab_others}94.7/11.9/25.8/59.5 &\cellcolor{tab_others}99.1/65.0/67.7/88.1 &\cellcolor{tab_ours}\textbf{99.7}/\textbf{83.8}/\textbf{77.9}/\textbf{94.1}\\

fryum & 97.0/47.6/51.5/93.4 & 95.9/34.0/40.6/76.2 &93.0/39.1/45.4/85.1 & \textbf{97.6}/\textbf{58.6}/\textbf{60.1}/81.3 &96.6/51.6/53.4/93.5 &\cellcolor{tab_ours}97.2/51.4/53.7/\textbf{96.2}\\

\cellcolor{tab_others}pipe\_fryum &\cellcolor{tab_others}99.1/56.8/58.8/95.4 & \cellcolor{tab_others}98.9/50.2/57.7/91.5 & \cellcolor{tab_others}98.5/65.6/63.4/83.0 &\cellcolor{tab_others}\textbf{99.4}/\textbf{72.7}/\textbf{69.9}/89.9 &\cellcolor{tab_others}99.2/64.3/65.1/95.2 &\cellcolor{tab_ours}99.3/63.7/68.4/\textbf{97.8}\\

\midrule
Mean & 98.1/38.0/42.6/91.8 & 95.9/21.0/27.0/75.6 & 96.8/34.7/37.8/81.4 &96.0/26.1/33.0/75.2 &98.7/53.2/55.7/94.5 &\cellcolor{tab_ours}\textbf{98.8}/\textbf{55.6}/\textbf{57.0}/\textbf{96.3}\\

\midrule
\cellcolor{tab_others}mAD &\cellcolor{tab_others}77.8  &\cellcolor{tab_others}74.6  &\cellcolor{tab_others}72.4  &\cellcolor{tab_others}70.1  & \cellcolor{tab_others}85.1 & \cellcolor{tab_ours}{\textbf{86.2}} \\

\bottomrule
\end{tabular}
}
\label{tab:visapx}%
\end{table}%
\begin{table}[!ht]
\centering
\caption{Per-class performance on \textbf{Real-IAD} dataset for multi-class anomaly detection with AUROC/AP/$F_1$-max metrics.}
\resizebox{1.\linewidth}{!}{
\begin{tabular}{cccccccc}
\toprule
Method~$\rightarrow$ & RD4AD~\cite{deng2022rd4ad} & UniAD~\cite{uniad} & SimpleNet~\cite{liu2023simplenet} & DiAD~\cite{he2024diad} & Dinomaly~\cite{guo2024dinomaly} & \cellcolor{tab_ours}CRR \\
\cline{1-1}
Category~$\downarrow$ & CVPR'22 & NeurlPS'22 & CVPR'23 & AAAI'24 & Arxiv'24 & \cellcolor{tab_ours} Ours \\
\hline
audiojack & 76.2/63.2/60.8 & 81.4/76.6/64.9 & 58.4/44.2/50.9 & 76.5/54.3/65.7 & 86.8/82.4/72.2 & \cellcolor{tab_ours}\textbf{90.2}/\textbf{85.4}/\textbf{76.3} \\

\cellcolor{tab_others}bottle cap & \cellcolor{tab_others}89.5/86.3/81.0 & \cellcolor{tab_others}92.5/91.7/81.7 & \cellcolor{tab_others}54.1/47.6/60.3 & \cellcolor{tab_others}91.6/\textbf{94.0}/\textbf{87.9} & \cellcolor{tab_others}\textbf{89.9}/86.7/81.2 & \cellcolor{tab_ours}93.5/92.0/83.3 \\

button battery & 73.3/78.9/76.1 & 75.9/81.6/76.3 & 52.5/60.5/72.4 & 80.5/71.3/70.6 & 86.6/88.9/\textbf{82.1} & \cellcolor{tab_ours}\textbf{87.4}/\textbf{90.3}/81.8 \\

\cellcolor{tab_others}end cap & \cellcolor{tab_others}79.8/84.0/77.8 & \cellcolor{tab_others}80.9/86.1/78.0 & \cellcolor{tab_others}51.6/60.8/72.9 & \cellcolor{tab_others}85.1/83.4/\textbf{84.8} & \cellcolor{tab_others}87.0/87.5/83.4 & \cellcolor{tab_ours}\textbf{89.1}/\textbf{88.9}/\textbf{85.7} \\

eraser & 90.0/88.7/79.7 & \textbf{90.3}/\textbf{89.2}/\textbf{80.2} & 46.4/39.1/55.8 & 80.0/80.0/77.3 & 90.3/87.6/78.6 & \cellcolor{tab_ours}\textbf{93.7}/92.6/83.7 \\

\cellcolor{tab_others}fire hood & \cellcolor{tab_others}78.3/70.1/64.5 & \cellcolor{tab_others}80.6/74.8/66.4 & \cellcolor{tab_others}58.1/41.9/54.4 & \cellcolor{tab_others}83.3/\textbf{81.7}/\textbf{80.5} & \cellcolor{tab_others}83.8/76.2/69.5 & \cellcolor{tab_ours}\textbf{86.5}/\textbf{80.6}/\textbf{73.0} \\

mint & 65.8/63.1/64.8 & 67.0/66.6/64.6 & 52.4/50.3/63.7 & \textbf{76.7}/\textbf{76.7}/\textbf{76.0} & 73.1/72.0/67.7 & \cellcolor{tab_ours}\textbf{77.1}/\textbf{77.0}/\textbf{69.5} \\

\cellcolor{tab_others}mounts & \cellcolor{tab_others}88.6/79.9/74.8 & \cellcolor{tab_others}87.6/77.3/77.2 & \cellcolor{tab_others}58.7/48.1/52.4 & \cellcolor{tab_others}75.3/74.5/\textbf{82.5} & \cellcolor{tab_others}\textbf{90.4}/\textbf{84.2}/78.0 & \cellcolor{tab_ours}89.6/81.2/78.0 \\

pcb & 79.5/85.8/79.7 & 81.0/88.2/79.1 & 54.5/66.0/75.5 &86.0/85.1/85.4 & 92.0/95.3/87.0 & \cellcolor{tab_ours}\textbf{93.0}/\textbf{95.7}/\textbf{88.4} \\

\cellcolor{tab_others}phone battery & \cellcolor{tab_others}87.5/83.3/77.1 & \cellcolor{tab_others}83.6/80.0/71.6 & \cellcolor{tab_others}51.6/43.8/58.0 & \cellcolor{tab_others}82.3/77.7/75.9 & \cellcolor{tab_others}92.9/91.6/82.5 & \cellcolor{tab_ours}\textbf{93.6}/\textbf{92.0}/\textbf{84.0} \\

plastic nut & 80.3/68.0/64.4 & 80.0/69.2/63.7 & 59.2/40.3/51.8 & 71.9/58.2/65.6 & 88.3/81.8/74.7 & \cellcolor{tab_ours}\textbf{91.5}/\textbf{87.5}/\textbf{78.4} \\

\cellcolor{tab_others}plastic plug & \cellcolor{tab_others}81.9/74.3/68.8 & \cellcolor{tab_others}81.4/75.9/67.6 & \cellcolor{tab_others}48.2/38.4/54.6 & \cellcolor{tab_others}88.7/\textbf{89.2}/\textbf{90.9} & \cellcolor{tab_others}90.5/86.4/78.6 & \cellcolor{tab_ours}\textbf{91.8}/\textbf{88.2}/\textbf{80.6} \\

porcelain doll & 86.3/76.3/71.5 & 85.1/75.2/69.3 & 66.3/54.5/52.1 & 72.6/66.8/65.2 & 85.1/73.3/69.6 & \cellcolor{tab_ours}\textbf{91.6}/\textbf{86.7}/\textbf{77.6} \\

\cellcolor{tab_others}regulator & \cellcolor{tab_others}66.9/48.8/47.7 & \cellcolor{tab_others}56.9/41.5/44.5 & \cellcolor{tab_others}50.5/29.0/43.9 & \cellcolor{tab_others}72.1/71.4/\textbf{78.2} & \cellcolor{tab_others}85.2/78.9/69.8 & \cellcolor{tab_ours}\textbf{88.9}/\textbf{84.6}/\textbf{77.2} \\

rolled strip base & 97.5/98.7/94.7 & 98.7/99.3/96.5 & 59.0/75.7/79.8 & 68.4/55.9/56.8 & 99.2/99.6/97.1 & \cellcolor{tab_ours}\textbf{99.3}/\textbf{99.7}/\textbf{97.4} \\

\cellcolor{tab_others}sim card set & \cellcolor{tab_others}91.6/91.8/84.8 & \cellcolor{tab_others}89.7/90.3/83.2 & \cellcolor{tab_others}63.1/69.7/70.8 & \cellcolor{tab_others}72.6/53.7/61.5 & \cellcolor{tab_others}95.8/96.3/88.8 & \cellcolor{tab_ours}\textbf{97.7}/\textbf{98.2}/92.2 \\

switch & 84.3/87.2/77.9 & 85.5/88.6/78.4 & 62.2/66.8/68.6 & 73.4/49.4/61.2 & \textbf{97.8}/\textbf{98.1}/\textbf{93.3} & \cellcolor{tab_ours}97.4/\textbf{97.9}/92.7 \\

\cellcolor{tab_others}tape & \cellcolor{tab_others}96.0/95.1/87.6 & \cellcolor{tab_others}\textbf{97.2}/\textbf{96.2}/\textbf{89.4} & \cellcolor{tab_others}49.9/41.1/54.5 & \cellcolor{tab_others}73.9/57.8/66.1 & \cellcolor{tab_others}96.9/95.0/88.8 & \cellcolor{tab_ours}\textbf{97.6}/\textbf{96.3}/\textbf{90.7} \\

terminalblock & 89.4/89.7/83.1 & 87.5/89.1/81.0 & 59.8/64.7/68.8 & 62.1/36.4/47.8 & 96.7/97.4/91.1 & \cellcolor{tab_ours}\textbf{97.0}/\textbf{97.6}/\textbf{91.4} \\

\cellcolor{tab_others}toothbrush & \cellcolor{tab_others}82.0/83.8/77.2 & \cellcolor{tab_others}78.4/80.1/75.6 & \cellcolor{tab_others}65.9/70.0/70.1 & \cellcolor{tab_others}\textbf{91.2}/\textbf{93.7}/\textbf{90.9} & \cellcolor{tab_others}90.4/91.9/83.4 & \cellcolor{tab_ours}\textbf{90.9}/\textbf{92.7}/\textbf{84.7}\\

toy & 69.4/74.2/75.9 & 68.4/75.1/74.8 & 57.8/64.4/73.4 & 66.2/57.3/59.8 & \textbf{85.6}/89.1/81.9 & \cellcolor{tab_ours}85.4/\textbf{88.3}/\textbf{81.9} \\

\cellcolor{tab_others}toy brick & \cellcolor{tab_others}63.6/56.1/59.0 & \cellcolor{tab_others}\textbf{77.0}/\textbf{71.1}/66.2 & \cellcolor{tab_others}58.3/49.7/58.2 & \cellcolor{tab_others}68.4/45.3/55.9 & \cellcolor{tab_others}72.3/65.1/63.4 & \cellcolor{tab_ours}\textbf{79.7}/\textbf{75.4}/\textbf{68.7} \\

transistor1 & 91.0/94.0/85.1 & 93.7/95.9/88.9 & 62.2/69.2/72.1 & 73.1/63.1/62.7 & \textbf{97.4}/\textbf{98.2}/\textbf{93.1} & \cellcolor{tab_ours}97.0/97.9/\textbf{93.1} \\

\cellcolor{tab_others}u block & \cellcolor{tab_others}89.5/85.0/74.2 & \cellcolor{tab_others}88.8/84.2/75.5 & \cellcolor{tab_others}62.4/48.4/51.8 & \cellcolor{tab_others}75.2/68.4/67.9 & \cellcolor{tab_others}89.9/\textbf{84.0}/\textbf{75.2} & \cellcolor{tab_ours}\textbf{93.8}/\textbf{91.3}/\textbf{82.8} \\

usb & 84.9/84.3/75.1 & 78.7/79.4/69.1 & 57.0/55.3/62.9 &58.9/37.4/45.7 & 92.0/91.6/83.3 & \cellcolor{tab_ours}\textbf{93.4}/\textbf{93.0}/\textbf{85.5} \\

\cellcolor{tab_others}usb adaptor & \cellcolor{tab_others}71.1/61.4/62.2 & \cellcolor{tab_others}76.8/71.3/64.9 & \cellcolor{tab_others}47.5/38.4/56.5 & \cellcolor{tab_others}76.9/60.2/67.2 & \cellcolor{tab_others}81.5/74.5/69.4 & \cellcolor{tab_ours}\textbf{85.6}/\textbf{82.2}/\textbf{72.5} \\

vcpill & 85.1/80.3/72.4 & 87.1/84.0/74.7 & 59.0/48.7/56.4 & 64.1/40.4/56.2 & 92.0/91.2/82.0 & \cellcolor{tab_ours}\textbf{93.7}/\textbf{92.7}/\textbf{84.3} \\

\cellcolor{tab_others}wooden beads & \cellcolor{tab_others}81.2/78.9/70.9 & \cellcolor{tab_others}78.4/77.2/67.8 & \cellcolor{tab_others}55.1/52.0/60.2 & \cellcolor{tab_others}62.1/56.4/65.9 & \cellcolor{tab_others}87.3/85.8/77.4 & \cellcolor{tab_ours}\textbf{89.9}/\textbf{88.4}/\textbf{79.6} \\

woodstick & 76.9/61.2/58.1 & 80.8/72.6/63.6 & 58.2/35.6/45.2 & 74.1/66.0/62.1 & 84.0/73.3/65.6 & \cellcolor{tab_ours}\textbf{85.2}/\textbf{76.4}/\textbf{68.1} \\

\cellcolor{tab_others}zipper & \cellcolor{tab_others}95.3/97.2/91.2 & \cellcolor{tab_others}98.2/98.9/95.3 & \cellcolor{tab_others}77.2/86.7/77.6 & \cellcolor{tab_others}86.0/87.0/84.0 & \cellcolor{tab_others}\textbf{99.1}/\textbf{99.5}/\textbf{96.5} & \cellcolor{tab_ours}98.7/99.3/95.9 \\

\midrule
Mean & 82.4/79.0/73.9 & 83.0/80.9/74.3 & 57.2/53.4/61.5 & 75.6/66.4/69.9 & 89.3/86.8/80.2 & \cellcolor{tab_ours}\textbf{91.3}/\textbf{89.7}/\textbf{82.6} \\
\bottomrule
\end{tabular}
}
\label{tab:realiadsp}
\end{table}
\begin{table}[!ht]
\centering
\caption{Per-class performance on \textbf{Real-IAD} dataset for multi-class anomaly localization with AUROC/AP/$F_1$-max/AUPRO metrics.}
\resizebox{1.\linewidth}{!}{
\begin{tabular}{cccccccc}
\toprule
Method~$\rightarrow$ & RD4AD~\cite{deng2022rd4ad} & UniAD~\cite{uniad} & SimpleNet~\cite{liu2023simplenet} & DiAD~\cite{he2024diad} & Dinomaly~\cite{guo2024dinomaly} & \cellcolor{tab_ours}CRR \\
\cline{1-1}
Category~$\downarrow$ & CVPR'22 & NeurlPS'22 & CVPR'23 & AAAI'24 & Arxiv'24 & \cellcolor{tab_ours}Ours \\
\hline
audiojack & 96.6/12.8/22.1/79.6 & 97.6/20.0/31.0/83.7 & 74.4/ 0.9/ 4.8/38.0 & 91.6/ 1.0/ 3.9/63.3 & 98.7/48.1/54.5/91.7 & \cellcolor{tab_ours}\textbf{99.4}/\textbf{65.0}/\textbf{63.9}/\textbf{94.0} \\

\cellcolor{tab_others}bottle cap & \cellcolor{tab_others}99.5/18.9/29.9/95.7 & \cellcolor{tab_others}99.5/19.4/29.6/96.0 & \cellcolor{tab_others}85.3/ 2.3/ 5.7/45.1 & \cellcolor{tab_others}94.6/ 4.9/11.4/73.0 & \cellcolor{tab_others}\textbf{99.7}/32.4/36.7/98.1 & \cellcolor{tab_ours}\textbf{99.9}/\textbf{47.9}/\textbf{47.7}/\textbf{98.9} \\

button battery& 97.6/33.8/37.8/86.5 & 96.7/28.5/34.4/77.5 & 75.9/ 3.2/ 6.6/40.5 & 84.1/ 1.4/ 5.3/66.9 & 99.1/46.9/56.7/92.9 & \cellcolor{tab_ours}\textbf{99.6}/\textbf{63.7}/\textbf{62.9}/\textbf{96.2} \\

\cellcolor{tab_others}end cap & \cellcolor{tab_others}96.7/12.5/22.5/89.2 & \cellcolor{tab_others}95.8/ 8.8/17.4/85.4 & \cellcolor{tab_others}63.1/ 0.5/ 2.8/25.7 & \cellcolor{tab_others}81.3/ 2.0/ 6.9/38.2 & \cellcolor{tab_others} 99.1/26.2/32.9/96.0 & \cellcolor{tab_ours}\textbf{99.5}/\textbf{32.2}/\textbf{38.8}/\textbf{97.6} \\

eraser & \textbf{99.5}/30.8/36.7/96.0 & 99.3/24.4/30.9/94.1 & 80.6/ 2.7/ 7.1/42.8 & 91.1/ 7.7/15.4/67.5 & 99.5/39.6/43.3/96.4 & \cellcolor{tab_ours}\textbf{99.8}/\textbf{52.8}/\textbf{54.1}/\textbf{98.5} \\

\cellcolor{tab_others}fire hood & \cellcolor{tab_others}98.9/27.7/35.2/87.9 & \cellcolor{tab_others}98.6/23.4/32.2/85.3 & \cellcolor{tab_others}70.5/ 0.3/ 2.2/25.3 & \cellcolor{tab_others}91.8/ 3.2/ 9.2/66.7 & \cellcolor{tab_others} 99.3/38.4/42.7/93.0 & \cellcolor{tab_ours}\textbf{99.6}/\textbf{47.7}/\textbf{47.8}/\textbf{93.9} \\

mint & 95.0/11.7/23.0/72.3 & 94.4/ 7.7/18.1/62.3 & 79.9/ 0.9/ 3.6/43.3 & 91.1/ 5.7/11.6/64.2 & 96.9/22.0/32.5/77.6 & \cellcolor{tab_ours}\textbf{98.6}/\textbf{28.7}/\textbf{36.2}/\textbf{85.6} \\

\cellcolor{tab_others}mounts & \cellcolor{tab_others}99.3/30.6/37.1/94.9 & \cellcolor{tab_others}\textbf{99.4}/28.0/32.8/95.2 & \cellcolor{tab_others}80.5/ 2.2/ 6.8/46.1 & \cellcolor{tab_others}84.3/ 0.4/ 1.1/48.8 & \cellcolor{tab_others} 99.4/39.9/44.3/95.6 & \cellcolor{tab_ours}\textbf{99.6}/\textbf{51.3}/\textbf{52.1}/\textbf{97.4} \\

pcb & 97.5/15.8/24.3/88.3 & 97.0/18.5/28.1/81.6 & 78.0/ 1.4/ 4.3/41.3 & 92.0/ 3.7/ 7.4/66.5 & 99.3/55.0/56.3/95.7 & \cellcolor{tab_ours}\textbf{99.5}/\textbf{66.1}/\textbf{62.7}/\textbf{96.7} \\

\cellcolor{tab_others}phone battery & \cellcolor{tab_others}77.3/22.6/31.7/94.5 & \cellcolor{tab_others}85.5/11.2/21.6/88.5 & \cellcolor{tab_others}43.4/ 0.1/ 0.9/11.8 & \cellcolor{tab_others}96.8/ 5.3/11.4/85.4 & \cellcolor{tab_others} 99.7/51.6/54.2/96.8 & \cellcolor{tab_ours}\textbf{99.9}/\textbf{69.4}/\textbf{63.6}/\textbf{98.0}\\

plastic nut& 98.8/21.1/29.6/91.0 &98.4/20.6/27.1/88.9 &77.4/ 0.6/ 3.6/41.5 &81.1/ 0.4/ 3.4/38.6 &99.7/41.0/45.0/97.4 & \cellcolor{tab_ours}\textbf{99.8}/\textbf{55.5}/\textbf{50.4}/\textbf{98.4}\\

\cellcolor{tab_others}plastic plug& \cellcolor{tab_others}99.1/20.5/28.4/94.9 &\cellcolor{tab_others}98.6/17.4/26.1/90.3 &\cellcolor{tab_others}78.6/ 0.7/ 1.9/38.8 &\cellcolor{tab_others}92.9/ 8.7/15.0/66.1 &\cellcolor{tab_others} 99.4/31.7/37.2/96.4 & \cellcolor{tab_ours}\textbf{99.7}/\textbf{39.9}/\textbf{44.9}/\textbf{97.9}\\

porcelain doll& 99.2/24.8/34.6/95.7 &98.7/14.1/24.5/93.2 &81.8/ 2.0/ 6.4/47.0 &93.1/ 1.4/ 4.8/70.4 &99.3/27.9/33.9/96.0 & \cellcolor{tab_ours}\textbf{99.7}/\textbf{36.3}/\textbf{40.5}/\textbf{98.0}\\

\cellcolor{tab_others}regulator& \cellcolor{tab_others}98.0/7.8/16.1/88.6 &\cellcolor{tab_others}95.5/9.1/17.4/76.1 &\cellcolor{tab_others}76.6/0.1/0.6/38.1 & \cellcolor{tab_others}84.2/0.4/1.5/44.4 & \cellcolor{tab_others} 99.3/42.2/48.9/95.6 & \cellcolor{tab_ours}\textbf{99.7}/\textbf{53.9}/\textbf{55.7}/\textbf{97.2}\\

rolled strip base& \textbf{99.7}/31.4/39.9/98.4 &99.6/20.7/32.2/97.8 &80.5/ 1.7/ 5.1/52.1 &87.7/ 0.6/ 3.2/63.4 &99.7/41.6/45.5/98.5 & \cellcolor{tab_ours}\textbf{99.9}/\textbf{58.3}/\textbf{56.3}/\textbf{99.3}\\

\cellcolor{tab_others}sim card set& \cellcolor{tab_others}98.5/40.2/44.2/89.5 &\cellcolor{tab_others}97.9/31.6/39.8/85.0 &\cellcolor{tab_others}71.0/ 6.8/14.3/30.8 &\cellcolor{tab_others}89.9/ 1.7/ 5.8/60.4 & \cellcolor{tab_others} 99.0/52.1/52.9/90.9 & \cellcolor{tab_ours}\textbf{99.7}/\textbf{77.6}/\textbf{70.4}/\textbf{96.5}\\

switch& 94.4/18.9/26.6/90.9 &98.1/33.8/40.6/90.7 &71.7/ 3.7/ 9.3/44.2 &90.5/ 1.4/ 5.3/64.2 &\textbf{96.7}/62.3/63.6/\textbf{95.9} & \cellcolor{tab_ours}95.9/\textbf{63.6}/\textbf{64.9}/95.6\\

\cellcolor{tab_others}tape& \cellcolor{tab_others}99.7/42.4/47.8/98.4 &\cellcolor{tab_others}99.7/29.2/36.9/97.5 &\cellcolor{tab_others}77.5/ 1.2/ 3.9/41.4 &\cellcolor{tab_others}81.7/ 0.4/ 2.7/47.3 &\cellcolor{tab_others} \textbf{99.8}/54.0/55.8/98.8 & \cellcolor{tab_ours}\textbf{99.9}/\textbf{65.4}/\textbf{62.6}/\textbf{99.1}\\

terminalblock& 99.5/27.4/35.8/97.6 &99.2/23.1/30.5/94.4 &87.0/ 0.8/ 3.6/54.8 &75.5/ 0.1/ 1.1/38.5 &\textbf{99.8}/48.0/50.7/98.8 & \cellcolor{tab_ours}\textbf{99.8}/\textbf{62.8}/\textbf{59.1}/\textbf{99.2}\\

\cellcolor{tab_others}toothbrush& \cellcolor{tab_others}96.9/26.1/34.2/88.7 &\cellcolor{tab_others}95.7/16.4/25.3/84.3 &\cellcolor{tab_others}84.7/ 7.2/14.8/52.6 &\cellcolor{tab_others}82.0/ 1.9/ 6.6/54.5 &\cellcolor{tab_others}96.9/\textbf{38.3}/\textbf{43.9}/\textbf{90.4} & \cellcolor{tab_ours}\textbf{97.5}/37.4/42.8/89.8\\

toy & 95.2/ 5.1/12.8/82.3 & 93.4/ 4.6/12.4/70.5 &67.7/ 0.1/ 0.4/25.0 &82.1/ 1.1/ 4.2/50.3 &\textbf{94.9}/22.5/32.1/\textbf{91.0} & \cellcolor{tab_ours}94.7/\textbf{32.5}/\textbf{40.2}/90.0\\

\cellcolor{tab_others}toy brick& \cellcolor{tab_others}96.4/16.0/24.6/75.3 &\cellcolor{tab_others}\textbf{97.4}/17.1/27.6/\textbf{81.3} &\cellcolor{tab_others}86.5/ 5.2/11.1/56.3 &\cellcolor{tab_others}93.5/ 3.1/ 8.1/66.4 &\cellcolor{tab_others}96.8/27.9/34.0/76.6 & \cellcolor{tab_ours}98.2/\textbf{43.8}/\textbf{48.2}/\textbf{85.2}\\

transistor1& 99.1/29.6/35.5/95.1 &98.9/25.6/33.2/94.3 &71.7/ 5.1/11.3/35.3 &88.6/ 7.2/15.3/58.1 &99.6/53.5/53.3/97.8 & \cellcolor{tab_ours}\textbf{99.6}/\textbf{63.2}/\textbf{59.2}/\textbf{98.0}\\

\cellcolor{tab_others}u block& \cellcolor{tab_others}99.6/40.5/45.2/96.9 &\cellcolor{tab_others}99.3/22.3/29.6/94.3 &\cellcolor{tab_others}76.2/ 4.8/12.2/34.0 &\cellcolor{tab_others}88.8/ 1.6/ 5.4/54.2 &\cellcolor{tab_others} \textbf{99.5}/41.8/45.6/96.8 & \cellcolor{tab_ours}\textbf{99.8}/\textbf{54.2}/\textbf{53.7}/\textbf{98.3}\\

usb& 98.1/26.4/35.2/91.0 &97.9/20.6/31.7/85.3 &81.1/ 1.5/ 4.9/52.4 &78.0/ 1.0/ 3.1/28.0 &\textbf{99.2}/45.0/48.7/97.5 & \cellcolor{tab_ours}\textbf{99.4}/\textbf{52.3}/\textbf{53.9}/\textbf{97.5}\\

\cellcolor{tab_others}usb adaptor& \cellcolor{tab_others}94.5/ 9.8/17.9/73.1 &\cellcolor{tab_others}96.6/10.5/19.0/78.4 &\cellcolor{tab_others}67.9/ 0.2/ 1.3/28.9 & \cellcolor{tab_others}94.0/ 2.3/ 6.6/75.5 &\cellcolor{tab_others} 98.7/23.7/32.7/91.0 & \cellcolor{tab_ours}\textbf{99.5}/36.2/39.2/\textbf{96.6}\\

vcpill& 98.3/43.1/48.6/88.7 &99.1/40.7/43.0/91.3 &68.2/ 1.1/ 3.3/22.0 &90.2/ 1.3/ 5.2/60.8 &99.1/66.4/66.7/93.7 & \cellcolor{tab_ours}\textbf{99.3}/\textbf{75.1}/\textbf{71.2}/\textbf{95.3}\\

\cellcolor{tab_others}wooden beads&	\cellcolor{tab_others}98.0/27.1/34.7/85.7 &\cellcolor{tab_others}97.6/16.5/23.6/84.6 &\cellcolor{tab_others}68.1/ \;2.4/ \;6.0/28.3 &\cellcolor{tab_others}85.0/ \;1.1/ \;4.7/45.6 &\cellcolor{tab_others}99.1/45.8/50.1/90.5 & \cellcolor{tab_ours}\textbf{99.6}/\textbf{60.2}/\textbf{59.2}/\textbf{94.7}\\

woodstick& 97.8/30.7/38.4/85.0 &94.0/36.2/44.3/77.2 &76.1/ 1.4/ 6.0/32.0 &90.9/ 2.6/ 8.0/60.7 &99.0/50.9/52.1/90.4 & \cellcolor{tab_ours}\textbf{99.4}/\textbf{68.4}/\textbf{65.4}/\textbf{92.8}\\

\cellcolor{tab_others}zipper& \cellcolor{tab_others}99.1/44.7/50.2/96.3 &\cellcolor{tab_others}98.4/32.5/36.1/95.1 &\cellcolor{tab_others}89.9/23.3/31.2/55.5 &\cellcolor{tab_others}90.2/12.5/18.8/53.5 &\cellcolor{tab_others}\textbf{99.3}/\textbf{67.2}/\textbf{66.5}/\textbf{97.8} & \cellcolor{tab_ours}99.2/57.7/58.1/97.7\\
\hline
Mean& 97.3/25.0/32.7/89.6 &97.3/21.1/29.2/86.7 &75.7/ 2.8/ 6.5/39.0 &88.0/ 2.9/ 7.1/58.1 &98.8/42.8/47.1/93.9 & \cellcolor{tab_ours}\textbf{99.2}/\textbf{54.0}/\textbf{54.2}/\textbf{95.8}\\

\hline
\cellcolor{tab_others}mAD &\cellcolor{tab_others}68.6  &\cellcolor{tab_others}67.5  &\cellcolor{tab_others}42.3  &\cellcolor{tab_others}52.6  & \cellcolor{tab_others}77.0 & \cellcolor{tab_ours}{\textbf{81.0}} \\

\bottomrule
\end{tabular}
}
\label{tab:realiadpx}
\end{table}
%

\begin{table}[!ht]
    \centering
    \small
    \caption{Image-level multi-class anomaly classification results with mAU-ROC/mAP/m$F_1$-max metrics on \textbf{HSS-IAD}.}
    \label{tab:hssiadsp}
    \renewcommand{\arraystretch}{1.1}
    \setlength\tabcolsep{3.0pt}
    \resizebox{1.0\linewidth}{!}{
        \begin{tabular}{p{0.2cm}<{\centering} p{2.0cm}<{\centering} p{2.0cm}<{\centering} p{2.0cm}<{\centering} p{2.0cm}<{\centering} p{2.0cm}<{\centering} p{2.0cm}<{\centering} p{2.0cm}<{\centering} p{3.6cm}<{\centering}}
            \toprule[1.5pt]
            \multicolumn{2}{c}{Method~$\rightarrow$} & DeSTSeg~\cite{zhang2023destseg}  & SimpleNet~\cite{liu2023simplenet} & DRAEM~\cite{zavrtanik2021draem} & UniAD~\cite{uniad} & RD4AD~\cite{deng2022rd4ad} & Dinomaly~\cite{guo2024dinomaly} & \cellcolor{tab_ours}{CRR} \\ 
            \cline{1-2}
            \multicolumn{2}{c}{Category~$\downarrow$} & CVPR'23 & CVPR'23 & ICCV'21 & NeuIPS'22 & CVPR'22   & CVPR'25 & \cellcolor{tab_ours}{(Ours)} \\
            \hline
            \multirow{5}{*}{\rotatebox{90}{Texture}} & MTD &71.2/89.2/88.0  &60.3/84.9/87.4  &50.7/79.2/87.4  &80.7/91.9/87.4 &86.2/95.7/89.7 & 93.6/98.3/93.9  & \cellcolor{tab_ours}91.2/97.5/92.3 \\
            
            & \cellcolor{tab_others}{STEEL} & \cellcolor{tab_others}{59.1/58.2/64.1} & \cellcolor{tab_others}{52.5/48.3/64.0} & \cellcolor{tab_others}{62.9/58.6/66.3} & \cellcolor{tab_others}{50.6/48.5/64.6} & \cellcolor{tab_others} 51.0/50.5/65.2 & \cellcolor{tab_others}{63.0/61.3/66.1}    & \cellcolor{tab_ours}58.1/59.3/65.6 \\
            
            & KolektorSDD &80.8/86.6/77.7  &60.1/63.2/70.9  &52.5/51.6/69.3  &66.1/82.3/77.7 & 80.9/82.2/78.1 & 93.0/93.5/88.5  & \cellcolor{tab_ours}93.5/93.4/87.6 \\
            
            & \cellcolor{tab_others}{KolektorSDD2} & \cellcolor{tab_others}{95.1/92.8/86.7} & \cellcolor{tab_others}{72.0/65.6/56.8} & \cellcolor{tab_others}{76.5/68.8/58.7} & \cellcolor{tab_others}{94.7/82.4/73.1} & \cellcolor{tab_others}94.8/91.6/85.2 & \cellcolor{tab_others}{93.3/91.5/85.4} & \cellcolor{tab_ours}92.0/89.7/99.1 \\
            \hline
            
            \multirow{3}{*}{\rotatebox{90}{Object}}& Casting\_C1 &69.7/75.0/82.6 &45.6/63.7/76.0  &66.2/69.6/82.6  &50.4/65.5/76.0 &51.3/70.5/76.0 & 71.5/83.3/80.9 & \cellcolor{tab_ours}74.1/85.4/79.2 \\
            
            & \cellcolor{tab_others}{Casting\_C2} & \cellcolor{tab_others}{63.0/63.8/74.2} & \cellcolor{tab_others}{55.2/63.0/73.1} & \cellcolor{tab_others}{67.3/\pzo0.6/76.4} & \cellcolor{tab_others}{50.5/63.9/72.6} & \cellcolor{tab_others}{56.7/64.0/71.6}& \cellcolor{tab_others}{61.6/63.9/74.2}   & \cellcolor{tab_ours}76.6/77.1/78.5 \\
            
            & Casting\_C3 &76.3/78.7/80.0  &34.5/46.9/71.9  &69.6/74.0/78.0 &50.5/66.2/73.3 & 63.3/65.1/75.0 & 67.6/67.9/78.6   & \cellcolor{tab_ours}77.1/80.6/80.7 \\
            \hline
            
            \multicolumn{2}{c}{\cellcolor{tab_others}Average} &\cellcolor{tab_others}73.6/77.8/79.0 &\cellcolor{tab_others}{54.3/62.2/71.4}  &\cellcolor{tab_others}{63.7/57.5/74.1} &\cellcolor{tab_others}{63.4/71.5/75.0}  & \cellcolor{tab_others}{69.2/74.2/77.3} & \cellcolor{tab_others}{77.7/79.9/81.1} & \cellcolor{tab_ours}\textbf{80.4}/\textbf{83.3}/\textbf{81.2} \\          
            \bottomrule[1.5pt]
        \end{tabular}
    }
\end{table}

\begin{table}[!ht]
    \centering
    \normalsize
    \caption{Pixel-level multi-class anomaly segmentation results (P-AUROC/P-AP/P-$F_1$-max/P-AUPRO) on \textbf{HSS-IAD}.}
    \renewcommand{\arraystretch}{1.1}
    \setlength\tabcolsep{1.0pt}
    \resizebox{1.0\linewidth}{!}{
        \begin{tabular}{p{0.25cm}<{\centering} p{2.5cm}<{\centering} p{3.0cm}<{\centering} p{3.0cm}<{\centering} p{3.0cm}<{\centering} p{3.0cm}<{\centering} p{3.0cm}<{\centering} p{3.0cm}<{\centering} p{3.5cm}<{\centering}}
            \toprule[1.5pt]
            \multicolumn{2}{c}{Method~$\rightarrow$} & DeSTSeg~\cite{zhang2023destseg} & SimpleNet~\cite{liu2023simplenet} & DRAEM~\cite{zavrtanik2021draem} & UniAD~\cite{uniad} & RD4AD~\cite{deng2022rd4ad} & Dinomaly~\cite{guo2024dinomaly}   & \cellcolor{tab_ours}CRR \\ 
            \cline{1-2}
            \multicolumn{2}{c}{Category~$\downarrow$} & CVPR'23 & CVPR'23 & ICCV'21 & NeurIPS'22 & CVPR'22 & CVPR'25    & \cellcolor{tab_ours}{(Ours)}  \\            
            \hline
            \multirow{5}{*}{\rotatebox{90}{Texture}} & MTD &72.0/26.4/28.5/70.6 &56.1/11.0/16.2/20.5  &59.1/11.7/18.0/20.6 &74.4/21.4/27.7/74.4 &  63.7/24.5/29.9/69.2 & 76.9/38.4/42.1/62.2    & \cellcolor{tab_ours}79.5/41.6/42.6/72.1 \\
            
            & \cellcolor{tab_others}{STEEL} & \cellcolor{tab_others}{74.3/17.9/23.1/36.0} & \cellcolor{tab_others}{57.0/\pzo3.9/\pzo8.7/\pzo9.6} & \cellcolor{tab_others}{64.9/13.8/19.0/18.3} & \cellcolor{tab_others}{79.5/19.3/26.3/37.6} & \cellcolor{tab_others}{75.6/18.6/24.2/48.9} & \cellcolor{tab_others}{84.0/30.7/32.7/43.3}    & \cellcolor{tab_ours}83.5/30.5/33.2/43.1 \\
            
            & KolektorSDD &85.9/17.5/27.5/33.8  &62.5/\pzo2.3/\pzo7.7/22.2  &67.9/\pzo0.2/\pzo0.6/\pzo6.0 &80.7/4.5/\pzo8.7/29.6 & 85.4/12.2/20.5/66.1 & 96.6/14.8/25.0/90.5    & \cellcolor{tab_ours}97.8/15.8/25.3/87.7 \\
            
            & \cellcolor{tab_others}{KolektorSDD2} & \cellcolor{tab_others}{97.4/69.5/66.0/91.8} & \cellcolor{tab_others}{92.8/49.0/53.3/78.7} & \cellcolor{tab_others}{82.2/29.7/35.3/46.2} & \cellcolor{tab_others}{97.6/46.2/49.1/94.1} & \cellcolor{tab_others}{97.2/52.0/54.4/92.6} & \cellcolor{tab_others}{98.5/68.3/65.1/87.6}  & \cellcolor{tab_ours}99.1/66.7/66.4/91.5 \\
            \hline
            
            \multirow{3}{*}{\rotatebox{90}{Object}}& Casting\_C1 &86.3/\pzo4.3/12.1/49.4  &37.8/\pzo0.1/\pzo0.3/\pzo2.7  &71.7/\pzo1.0/\pzo3.6/22.8  &74.8/\pzo0.9/\pzo3.0/31.5 &  80.2/\pzo3.6/\pzo8.5/43.3 & 80.1/\pzo2.6/\pzo7.7/40.2    & \cellcolor{tab_ours}88.6/\pzo5.7/13.2/55.1 \\
            
            & \cellcolor{tab_others}{Casting\_C2} & \cellcolor{tab_others}{85.7/\pzo1.5/\pzo4.0/54.8} & \cellcolor{tab_others}{32.9/\pzo0.1/\pzo0.2/\pzo1.4} & \cellcolor{tab_others}{74.1/\pzo0.6/\pzo2.1/29.2} & \cellcolor{tab_others}{85.0/\pzo1.2/\pzo4.0/47.8} & \cellcolor{tab_others}{83.8/\pzo2.0/\pzo5.3/56.1} & \cellcolor{tab_others}{82.7/\pzo2.1/\pzo6.0/45.2}    & \cellcolor{tab_ours}93.6/\pzo6.9/15.1/70.3 \\
            
            & Casting\_C3 &86.2/\pzo1.2/\pzo3.2/53.1 &40.0/\pzo0.1/\pzo0.3/\pzo8.5  &63.6/\pzo0.3/\pzo1.0/25.7 &72.6/\pzo0.4/\pzo1.4/32.1 & 73.4/\pzo1.0/\pzo3.1/30.2 & 68.0/\pzo0.4/\pzo1.0/14.9   & \cellcolor{tab_ours}79.9/\pzo2.1/\pzo7.6/34.8 \\
            \hline
            
            \multicolumn{2}{c}{\cellcolor{tab_others}Average} &\cellcolor{tab_others}84.0/19.8/23.5/55.6 &\cellcolor{tab_others}54.2/\pzo9.5/12.4/20.5  &\cellcolor{tab_others}70.5/\pzo8.2/11.4/24.1 &\cellcolor{tab_others}80.7/13.4/17.2/49.6 &\cellcolor{tab_others}79.9/16.3/20.8/58.1 & \cellcolor{tab_others}{83.8/22.5/25.7/54.8}   & \cellcolor{tab_ours}\textbf{88.8}/\textbf{24.2}/\textbf{29.1}/\textbf{64.9} \\
            \hline
            
            \multicolumn{2}{c}{mAD} &59.0  &40.6  &44.2  &53.0  & 56.5 & 60.8   & \cellcolor{tab_ours}\textbf{64.6} \\
            
            \bottomrule[1.5pt]
        \end{tabular}
    }
    \label{tab:hssiad_px}
\end{table}

